%% file: cas-dc-sample.tex
\documentclass[a4paper,fleqn]{cas-dc}

\usepackage[numbers]{natbib}
\usepackage{times}
\usepackage{graphicx}
\usepackage{caption}
\usepackage{subcaption}
\usepackage{listings}
\usepackage{latexsym}
\usepackage{todonotes}
\usepackage{float}
\usepackage{amsmath}
\usepackage{color}
\usepackage{enumitem}
\usepackage{cuted}
\def\tsc#1{\csdef{#1}{\textsc{\lowercase{#1}}\xspace}}
\tsc{WGM}
\tsc{QE}
\tsc{EP}
\tsc{PMS}
\tsc{BEC}
\tsc{DE}

\begin{document}
\let\WriteBookmarks\relax
\def\floatpagepagefraction{1}
\def\textpagefraction{.001}
\shorttitle{Monitoring Transformative Technological Convergence}
\shortauthors{Sternfeld, A. et al.}
\title [mode = title]{Monitoring Transformative Technological Convergence Through LLM-Extracted Semantic Entity Triple Graphs}       

\author[1]{Alexander Sternfeld}[
                        orcid=0009-0008-6801-6160]
\cormark[1]
\ead{alexander.sternfeld@hevs.ch}

\affiliation[1]{organization={Institute of Entrepeneurship and Management, HES-SO}, 
                city={Sierre},
                country={Switzerland}}

\author[1]{Andrei Kucharavy}[orcid=0000-0003-0429-8644]

\author[1]{Dimitri Percia David}[orcid=0000-0003-0429-8644]

\author[2]{Alain Mermoud}[orcid=0000-0001-6471-772X]

\author[2]{Julian Jang-Jaccard}[orcid=0000-0002-1002-057X]

\author[2]{Nathan Monnet}[]

\affiliation[2]{organization={Cyber-Defence Campus, armasuisse, Science and Technology},
                city={Thun},
                country={Switzerland}}

\cortext[cor1]{Corresponding author}

\begin{abstract}
Forecasting transformative technologies remains a critical but challenging task, particularly in fast-evolving domains such as Information and Communication Technologies (ICTs). Traditional expert-based methods struggle to keep pace with short innovation cycles and ambiguous early-stage terminology. In this work, we propose a novel, data-driven pipeline to monitor the emergence of transformative technologies by identifying patterns of technological convergence.

Our approach leverages advances in Large Language Models (LLMs) to extract semantic triples from unstructured text and construct a large-scale graph of technology-related entities and relations. We introduce a new method for grouping semantically similar technology terms (\emph{noun stapling}) and develop graph-based metrics to detect convergence signals. The pipeline includes multi-stage filtering, domain-specific keyword clustering, and a temporal trend analysis of topic co-occurence. 

We validate our methodology on two complementary datasets: 278{,}625 arXiv preprints (2017--2024) to capture early scientific signals, and 9{,}793 USPTO patent applications (2018--2024) to track downstream commercial developments. Our results demonstrate that the proposed pipeline can identify both established and emerging convergence patterns, offering a scalable and generalizable framework for technology forecasting grounded in full-text analysis.
\end{abstract}




\maketitle

\input{sections/introduction}
\input{sections/related_work}
\input{sections/data}

\input{sections/method}

\input{sections/results}

\input{sections/conclusion}




\bibliographystyle{cas-model2-names}

\bibliography{cas-refs}

\appendix
\input{sections/appendix}

\end{document}

%% file: sections/introduction.tex
\section{Introduction}
Accurate forecasting of technological disruptions is essential for organizations seeking to remain competitive and resilient. While incremental innovations can typically be anticipated and managed within existing planning frameworks, \textit{transformative technologies} introduce paradigm shifts that reshape entire domains and often spill over into adjacent sectors~\citep{Schumpeter1949, Ettlie1984OrganizationSA}. Because such technologies challenge established conceptual frameworks, they are inherently difficult to predict~\citep{Christensen1997InnovatorsDilemma}.

Despite these challenges, the strategic importance of anticipating transformative change has long motivated the development of forecasting methods, particularly since the Cold War era~\citep{TF1967Jantsch}. Seminal approaches such as the Delphi Method~\citep{dalkey1969experimental} and S-curve Substitution Analysis~\citep{fisher1971simple} have inspired decades of methodological refinement~\citep{national2010persistent, calleja_sanz2020, technology2004technology}. However, these classical methods struggle to keep pace with the rapid innovation cycles characteristic of Information and Communication Technologies (ICTs). For example, the term “Large Language Models” (LLMs) emerged in 2019, and within just three years, ChatGPT had become a globally recognized application~\citep{kucharavy2024deep}. In such contexts, the reliance of classical methods on expert panels and historical data renders them ill-suited~\citep{daim_digital_2022}.

As a result, ICT forecasting has increasingly shifted toward data-driven methodologies~\citep{daim_anticipating_2016}, with scientometrics emerging as a promising approach~\citep{perciadavid2023}. Yet even data-centric methods face difficulties in the early identification of transformative technologies, due in part to the absence of stable terminology and well-defined application domains in the initial stages. Continuing the example of LLMs, even a year after the release of ChatGPT, major bibliometric platforms such as OpenAlex still lacked ontology terms for many of the foundational sub-technologies~\citep{VelezEstevez2023NewTI, SinghChawla2022MassiveOI, würsch2023llms}. While \citet{würsch2023llms} showed that extracting technology-related nouns from scientific texts is feasible, the process remains noisy and challenging to use effectively for large-scale monitoring and forecasting.

In this work, we present an alternative approach that leverages recent advances in Large Language Models (LLMs) to enable scalable extraction and analysis of semantic triples centered on technological concepts~\citep{sternfeld_eeke}. We focus on \textit{technological convergence} as an early indicator of transformative potential. First introduced by \citet{rosenberg1963} in his study of the machine tool industry, convergence describes how formerly separate fields of science and technology come to rely on a shared technical and knowledge base, blurring the boundaries between them. Transformative technologies frequently arise at these points of overlap, so tracking where distinct fields begin to intersect offers an early signal of disruptive potential~\citep{Dosi1982TechnologicalPA, Li2024ANI}.


Our main contributions are as follows:
\begin{enumerate}[label=(\roman*)]
    \item We introduce an unbiased pipeline for extracting semantic triples that involve technology-designating nouns.
    \item We propose a novel syntactic similarity identification technique---termed \textit{noun stapling}---to group related technology terms.
    \item We implement a coarse decomposition of technology components to improve interpretability.
    \item We develop a graph-based method to detect and track technological convergence over time.
\end{enumerate}

These elements are integrated into a unified pipeline for identifying the emergence of transformative technologies through convergence patterns. We demonstrate the utility of this pipeline through a case study on LLMs, applying it to a large corpus of 278{,}625 arXiv preprints (2017--2024) and 9{,}793 USPTO patent applications (2018--2024).

The remainder of this paper is structured as follows: Section~\ref{Related work} reviews the related literature. Section~\ref{data} describes the data sources. Section~\ref{Methodology} outlines the methodology. Section~\ref{Results} presents the results. Finally, Section~\ref{Conclusion} concludes the paper and discusses directions for future research.

%% file: sections/related_work.tex
\section{Related work} \label{Related work}
In this section, three branches of previous research are discussed. First, we discuss the main topics and trends in bibliometrics. Then, we discuss related work on claim and triple extraction, where we highlight those methods that leverage machine learning techniques. Last, we discuss the usage of topological analyses for technology forecasting.

\subsection{Bibliometrics}

As a method for technological forecasting, bibliometrics leverages both qualitative and quantitative analysis of recorded information—primarily scientific books, articles, and patents. The systematic study of written works has long been established, with the term \emph{bibliometrics} coined in the early 20\textsuperscript{th} century. One of the most widely recognized classical bibliometric techniques is citation analysis, which examines citation patterns among academic publications~\citep{smith1981citation}.

Early citation analysis focused primarily on citation counts and co-citation networks. However, the advent of advanced network analysis methods and increasing computational power has enabled more sophisticated analyses of citation networks. Such methods are now employed across a range of fields for technology forecasting. For instance, \citet{qiu2022} conducted a citation network analysis of robotics patents to identify key pathways of innovation. Similarly, \citet{LI2020} analyzed both patents and academic papers in the domain of nanogenerator technologies, offering a more integrated view of scientific and technological progress.

Traditionally, bibliometric studies were constrained to structured metadata. Recent advances in Natural Language Processing (NLP), however, have opened new avenues for deeper analysis of full-text documents. Techniques such as $n$-gram analysis~\citep{michel2011quantitative}, Latent Dirichlet Allocation (LDA)~\citep{LDA2003}, semantic embeddings~\citep{Mikolov2013DistributedRO}, deep neural language model encodings~\citep{Context2Vec2016}, and, more recently, large language models (LLMs)~\citep{Beltagy2019SciBERT}, have greatly expanded the scope of bibliometric inquiry. These developments are further supported by the growing availability of full-text documents through open-access and preprint platforms, such as arXiv\footnote{\url{https://info.arxiv.org/}}, medXiv~\footnote{\url{https://www.medrxiv.org/content/about-medrxiv}}, bioRxiv\footnote{\url{https://www.biorxiv.org/content/about-biorxiv}}, and PubMed Central\footnote{\url{https://pmc.ncbi.nlm.nih.gov/about/intro/}}.

For example, \citet{perciadavid2023} analyzed computer science preprints on arXiv using LLM-derived sentence embeddings to extract author sentiment and relate it to technology security over time. Likewise, Sandu et al.~(2024) constructed a collaboration network in the area of social media text mining and identified recurring research themes using $n$-gram analysis.

In parallel with developments in literature-based bibliometrics, there has been increasing attention to patent analysis, given the rich technical detail and close ties to commercial applications that patents offer. While academic papers often represent early-stage scientific exploration, patents typically reflect efforts to translate these discoveries into deployable technologies. Importantly, both European and U.S. patents follow the Cooperative Patent Classification (CPC) system, enabling systematic, cross-jurisdictional analysis. For example, \citet{KIM2017228} cluster patents based on CPC codes and apply citation and claim analysis to forecast emerging technologies. Similarly, \citet{you2017} analyze light generator technology patents using Bass and ARIMA models to identify promising innovation trajectories. By integrating bibliometric insights from both scholarly publications and patents, researchers gain a more comprehensive view of technological development—spanning foundational research to commercial realization~\citep{LI2020, ADAMUTHE2019181}.

\subsection{Claim and triple extraction}

Semantic Triples, also known as RDF (Resource Description Framework) triples, are a fundamental representation for knowledge extraction and organization in semantic web technologies. A semantic triple consists of three components: a subject, a predicate, and an object, and is used to represent factual statements in a machine-readable format~\cite{berners2001rdf}. This structure is foundational in knowledge graphs, enabling systems to store and reason about facts in a way that can be queried and expanded upon~\cite{Hogan_2021}. In the context of claim extraction from scientific literature, we aim to extract claims that can be represented as semantic triples, focusing on technological assertions and relationships. By adhering to the semantic triple standard, these claims are organized into unrooted, unbiased knowledge graphs that facilitate downstream tasks such as pattern mining and knowledge discovery.

Previous approaches to claim extraction can be broadly categorized into heuristic and machine learning methods. Heuristic methods, such as the approach by ~\citet{jansen2017}, do not require training data and have low computational cost. However, they are limited in their ability to capture complex claims. \citet{jansen2017} extract core claims by scoring sentences based on pattern matching, term frequency, and sentence length. However, they only succeed in rewriting claims to the required AIDA (atomic, independent, declarative, and absolute) standard in only 29\% of cases. In contrast, machine learning methods like the one proposed by ~\citet{li2021}, using BiLSTMs for sentence classification, offer better performance, categorizing sentences with an F1 score over 0.8. However, these models require domain-specific supervised training data, which must be updated regularly.

Most triple extraction models rely on supervised training, such as RECON and sPERT, which require labeled training data~\cite{bastos2021, eberts2020}. These methods are limited by their dependence on available training data and predefined relations. A long-standing research stream is Open Information Extraction (OIE), first framed by \citet{openie} as the schema-free, domain-independent extraction of relational tuples from text. Early OIE systems relied on shallow syntactic patterns (TextRunner~\citep{textrunner}, ReVerb~\citep{reverb}), later incorporating deeper linguistic features (OLLIE~\citep{ollie}, ClausIE~\citep{clausie}) and clause-based reasoning, of which Stanford OpenIE~\citep{stanfordie} is a widely used representative. The recurring limitation across these generations is that schema-free extraction favours coverage over precision, so systems tend to over-generate surface-level or non-useful tuples~\citep{liu2022}, which motivates approaches that constrain extraction without reverting to full supervision. \citet{ottersen2024} address this issue by using a language model to extract triples within a predefined set of relations, improving precision but requiring users to specify all possible relations. 


More recently, extraction has shifted toward \emph{generative} formulations in which a language model emits triples directly as text. REBEL~\citep{rebel} casts relation extraction as end-to-end sequence generation, while UIE~\citep{uie} unifies multiple extraction tasks through a single text-to-structure schema. With instruction-tuned LLMs, this line has moved further toward few-shot and zero-shot extraction driven by prompting rather than task-specific training. ChatIE~\citep{chatie} decomposes zero-shot extraction into a multi-turn question-answering dialogue, and GPT-RE~\citep{gptre} uses in-context demonstrations to improve relation extraction. A parallel concern is enforcing valid output structure: constrained or structured decoding restricts generation so that outputs conform to the target format~\citep{uie, li2024}. For knowledge-graph construction specifically, EDC~\citep{edc} combines open extraction with schema definition and canonicalisation, showing that LLMs can produce high-quality triples without parameter tuning. The suitability of LLM-based extraction is domain-dependent; \citet{würsch2023llms} report that LLMs extract poorly matched knowledge entities from cybersecurity literature, which underscores that generative extraction must be paired with validation and filtering. A comprehensive survey of this generative IE paradigm, including prompt design, zero-shot learning, and constrained decoding, is provided by \citet{xu2023}.


Our work performs schema-free triple extraction from full-text scientific articles and patents using few-shot prompting of LLMs, thereby retaining the domain independence of OIE while exploiting the contextual capacity of LLMs to curb its characteristic over-generation. Unlike generative IE methods evaluated on sentence-level benchmarks~\citep{rebel, uie, chatie, gptre, xu2023}, we apply open extraction at the scale of hundreds of thousands of documents to build large, unbiased technology graphs.



\subsection{Topological Analyses for Technology Forecasting}

Traditional co-word analysis techniques primarily relied on co-occurrence frequencies to detect associations between terms~\citep{callon1983coword}. More recent approaches, however, emphasize the connectivity and centrality of entities within complex graphs to identify zones of technological convergence and emerging innovation clusters. At the core of these methods are \textit{knowledge graphs}, which are constructed by extracting key entities such as technologies, methods, and applications from scientific publications and patents. In these graphs, nodes represent concepts, and edges encode semantic, citation, or temporal relationships. Keyword extraction tools, such as RAKE~\cite{rake} and KeyBERT, are often used to enrich these graphs. The resulting structures can be analyzed as either static or dynamic representations of technological domains.

The topological characteristics of knowledge graphs are crucial for forecasting applications. For instance, \citet{DOTSIKA2017114} demonstrated that keyword co-occurrence networks derived from scientific and business literature can be effectively analyzed using centrality metrics to detect potentially disruptive or fast-growing technological areas. Building on this idea, \citet{LI2020} developed dual-layer networks that integrate scientific publications and patents to trace the evolution and commercialization of nanogenerator technologies. More recently, \citet{WANG2025101606} applied semantic term extraction in combination with Louvain-based community detection to identify early-stage convergence between disparate research fields.

Community detection continues to be a foundational component of topological forecasting. The Louvain algorithm, introduced by \citet{blondel2008LouvainMethod}, is widely used to identify coherent topic clusters within large-scale graphs. Complementary measures, such as eigenvector centrality, highlight influential nodes that play a key role in the dissemination and integration of knowledge. These techniques support dynamic analyses of how scientific topics emerge, grow, merge, or decline over time. Dynamic knowledge graphs, such as those implemented in the Science4Cast project~\citep{krenn2023science4cast}, enable the longitudinal monitoring of research landscapes and facilitate the detection of novel knowledge pathways as they form.

In addition, predictive modeling on these graphs—using either neural networks or handcrafted topological features—can forecast whether previously unconnected concepts are likely to become linked in the future, and whether such connections are poised to attract significant attention~\citep{gu2024forecasting, martinez2016survey}.

%% file: sections/data.tex
\section{Data} \label{data}
The methodology developed in this work can be used for any textual data source. Several examples of such sources are scientific articles, patents, news articles, social media posts and web pages. In this paper, we focus on two data sources: scientific articles and US patent applications. In order to evaluate the capabilities of different methods of technology-related semantic triples extraction, we created a golden dataset of manually labeled semantic triples in the LLM field.

\subsection{Golden Dataset for Triples Extraction}

To fine-tune the LLMs for technology-related semantic triples extraction, a labeled dataset is required, both for evaluation of method performance, and for supervised training. To the best of our knowledge, such a labeled dataset for semantic triples extraction from scientific papers does not exist. Therefore, we manually construct a training dataset based on the paper \textit{A Survey of Large Language Models} ~\cite{zhao2023survey}. We chose this paper given our focus on the LLMs transformative technology, since it is a comprehensive and general review of LLM component technologies.
In total, we manually annotated 547 triples in 100 text segments of 15 lines, given that amount of examples is generally considered to be sufficient for parameter-efficient fine-tuning (PEFT), while leaving room for a test holdout set~\cite{falissard-etal-2023-improving}. To evaluate the performance of the fine-tuned LLMs, we use a holdout set of 20\% of the annotated paragraphs. The dataset is available for download from the data and code repository associated to this article: \href{https://github.com/submissiontfscanonymized/tfsc2025}{https://github.com/submissiontfscanonymized/tfsc2025}.

\subsection{arXiv data}
In the first use case, preprints publicly available on arXiv are considered. As arXiv is an open scholarly preprint archive, the documents are not peer-reviewed. We choose this platform for two key reasons. First, as there is no peer review and only a short moderation process, submissions to arXiv are available rapidly, tracking the state of research as close to real time as possible. The archive includes papers that may not have been accepted at conferences due to perceived lack of immediate interest, yet have gone on to receive significant citations — highlighting their eventual importance. Second, arXiv ensures that all articles are classified by the authors into a category from a pre-defined taxonomy. To aid authors, arXiv has published an algorithm that can suggest the best fitting categories for a paper, based on the existing corpus of articles \footnote{https://info.arxiv.org/help/api/classify.html\#cat}. Finally, all arXiv pdf's and raw latex files can be downloaded through Google Cloud Storage Buckets, which are updated weekly \footnote{https://www.kaggle.com/datasets/Cornell-University/arxiv}, making text mining on arXiv preprings significantly easier.

For an initial evaluation and refinement of our methodology, we only worked with papers from December 2023 (4225 papers), from the arXiv categories \verb|cs.AI|, \verb|cs.CL| and \verb|cs.LG|. We choose to do so, as a small dataset reduces the computation time and required computational resources. Therefore, such a dataset is suitable for early refinements of the pipeline. After settling on the final methodology, all papers between 2018 and 2024 are considered, from the arXiv categories that are specified in Table \ref{tab:arxiv_cats} for a total of  278,625 articles. Both the categories and timeframe correspond to the timeline of convergence of technologies that led to the transformative LLM technology emergence.

\begin{table}[h]
\centering
\caption{The arXiv categories that are considered for this study, alongside their full names.}
\label{tab:arxiv_cats}
\begin{tabular}{@{}ll@{}}
\toprule
\textbf{Category} & \textbf{Full name}       \\ \midrule
cs.CL             & Computation and Language \\
cs.LG             & Machine Learning         \\
cs.AI             & Artificial Intelligence  \\
cs.IR             & Information Retrieval    \\
stat.ML           & Machine Learning         \\ \bottomrule
\end{tabular}%
\end{table}

\subsection{USPTO patent data}
The second data source we consider is USPTO patent data. We choose to consider patents as a secondary data source as it provides information on downstream applications of technologies. While preprints on arXiv will show early signs of novel technological developments, the presence of technologies in patent applications shows that technologies are maturing and are being adopted in commercial products.

Specifically, we consider patent applications to the USPTO, which are publicly available through the Bulk Data Directory \footnote{https://data.uspto.gov/bulkdata/datasets}. Similar to the arXiv papers, we consider the period 2018-2024. All patent applications are categorized according to the Cooperative Patent Classification (CPC) system, which is a classification system that is used both by te USPTO and the European Patent Office (EPO). 

For our case study, we filter the data for those patents that contain at least one of the terms \textit{large language model}, \textit{large language models}, \textit{llm} or \textit{llms} in the abstract or title. This results in a total of 9793 patent applications between 2018 and 2024. 


%% file: sections/method.tex
\section{Methodology} \label{Methodology}
\begin{figure*}[b] 
    \centering
    \includegraphics[width=\textwidth]{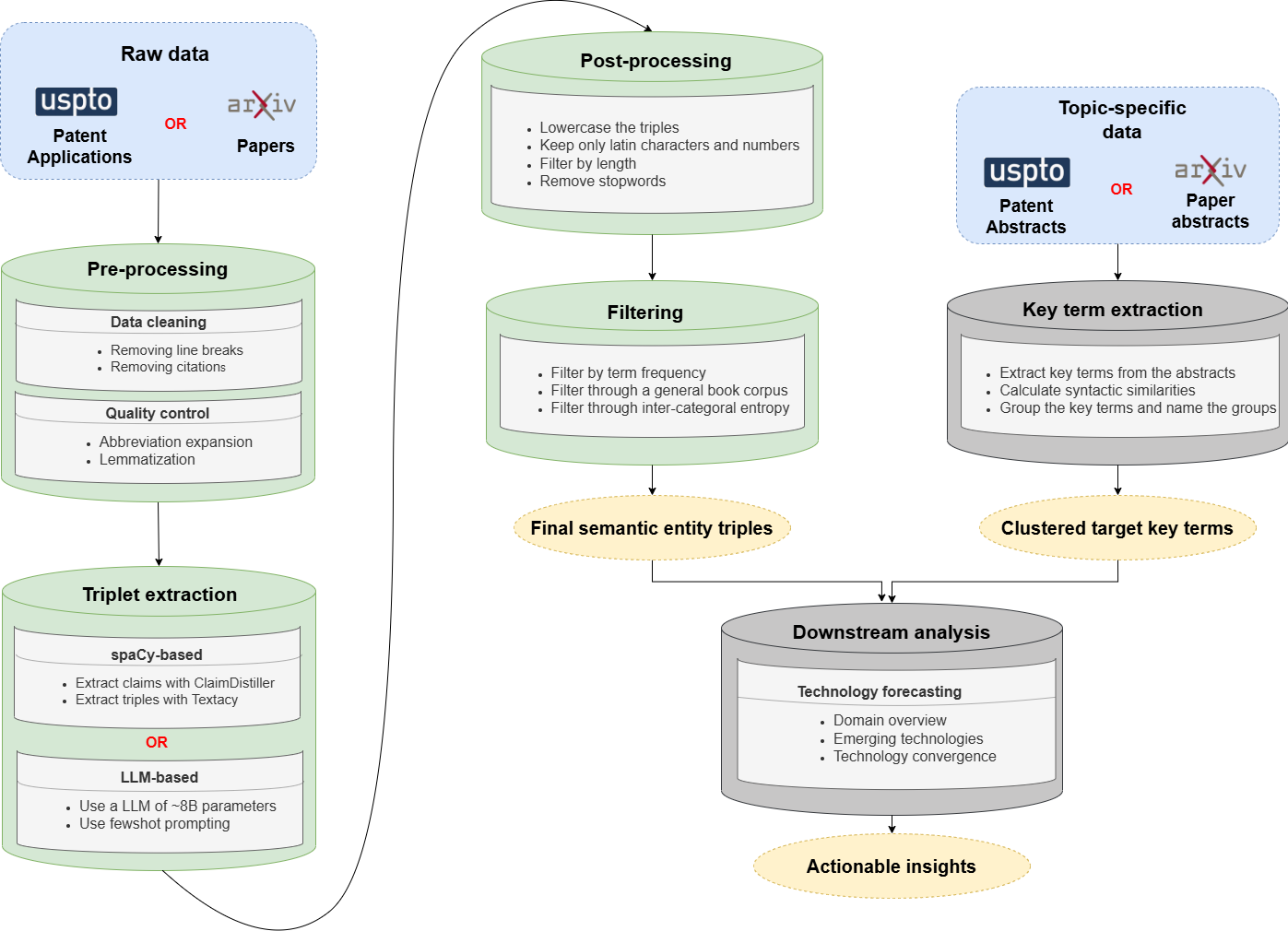}
    \caption{The complete pipeline for the triple extraction and downstream analysis, starting from either raw patent data or raw arXiv data. The green components of the pipeline reflect the triple extraction procedure, including the pre- and post-processing steps. The grey components of the pipeline illustrate the key-term extraction, noun stapling and the downstream analyses.}
    \label{fig:pipeline}
\end{figure*}

In this Section we describe the methodology of our triple extraction pipeline and downstream analyses. We first describe the data preprocessing, after which two triple extraction methods are presented. Then, the post-processing and filtering of the triples is elaborated upon. Last, we explain our proposed \emph{noun stapling} methods to group similar triples, and our downstream analyses. The full pipeline is displayed in Figure \ref{fig:pipeline}.

\subsection{Preprocessing}
The \verb|PyMuPDF| pdf processing library \cite{PyMuPDF} is used to turn the pdfs into text. The resulting text is then processed to remove citations, using a heuristic pattern matching. We remove both square and round brackets with only numbers inside, as well as brackets with a year inside, matching most frequent citation formats found in the arXiv papers. In addition to that, we remove the line breaks using a similar heuristic, although with dashes.

In order to unify extracted entities, it is important to expand abbreviations in the text. By doing so, we will be able to later on map \textit{LLM} and \textit{Large Language Model} to the same entity. To ensure that the abbreviation resolution is done properly, we benchmark different algorithms on the \textit{Fundamentals of Generative Large Language Models
and Perspectives in Cyber-Defense} report, given its mixed usage of both ML, cybersecurity, and cyberdefense abbreviations~\cite{kucharavy2023fundamentals}. Table \ref{tab:abbr} shows the comparative performance of the Schwartz-Hearst ~\cite{Schwartz2003}, scispaCy ~\cite{neumann-etal-2019-scispacy}, NLPRe ~\cite{NLPRe} and a fine-tuned RoBERTa model ~\cite{zilio-etal-2022-plod} as abbreviation detection methods. The results show that the Schwartz-Hearst algorithm performs the best, with scispaCy coming a close second. Given the speed of Schwartz-Hearst, we select it for our pipeline. Table \ref{tab:preprocessing} shows an example of a sentence before and after preprocessing.

\begin{table*}[!htb]
\centering
\caption{Performance of three abbreviation detection algorithms on the \textit{Fundamentals of Generative Large Language Models
and Perspectives in Cyber-Defense~\cite{kucharavy2023fundamentals}}. Manual annotation resulted in 15 abbreviations being explicitly introduced in the report, on which the percentages are based. }
\label{tab:abbr}
\begin{tabular}{@{}lccl@{}}
\toprule
                   & \multicolumn{1}{l}{Correctly detected} & \multicolumn{1}{l}{False positive} & Time   \\ \midrule
Schwartz-Hearst    & 11 (73\%)                                     & 1                                  & 0.04 s \\
scispaCy           & 11 (73\%)                                   & 4                                  & 8.73 s \\
NLPRe              & 0 (0\%)                                      & 0                                  & 0.01 s \\
Fine-tuned RoBERTa & 10 (67\%)                                     & 3                                  & 100 s  \\ \bottomrule
\end{tabular}%
\end{table*}

\begin{table*}[!t]
\caption{Illustration of the preprocessing effect, the parts in bold are altered during preprocessing.}
\centering
\label{tab:preprocessing}
\begin{tabular}{@{}ll@{}}
\toprule
Uncleaned                                                                                                                                            & Preprocessed                                                                                                                                                                                  \\ \midrule
\begin{tabular}[c]{@{}l@{}}Society has been affected by \\ \textbf{artificial intelligence (AI)} and has\\ become more \textbf{rel- iant} on \textbf{AI} products.\end{tabular} & \begin{tabular}[c]{@{}l@{}}Society has been affected by \\ \textbf{artificial intelligence} and has become\\  more \textbf{reliant} on \textbf{artificial intelligence} products.\end{tabular} \\ \bottomrule
\end{tabular}%
\end{table*}

\subsection{Triple extraction}

\subsubsection{spaCy-based triple extraction}
\paragraph{Claim extraction.}
After preprocessing the raw text, the sentences that constitute the core claims of the text are identified. To this end, the \textit{ClaimDistiller} framework is used, which is developed by \citet{Wei2023}. In this work, both CNN and BiLSTM models have been employed for claim extraction tasks through training on the PubMED-RCT and SciARK datasets \cite{fergadis-etal-2021-argumentation, dernoncourt-lee-2017-pubmed}. While incorporating supervised contrastive learning has been shown to enhance model performance, it also introduces additional computational cost. To maintain a balance between effectiveness and efficiency, we opt for the BiLSTM model variant without supervised contrastive training. This model is used to extract claims from academic papers, thereby reducing the number of sentences that need to be processed during the subsequent triple extraction phase.

\paragraph{Triple extraction.}
Next, we aim to reduce the claims to (\textit{subject, predicate, object}) semantic triples, analogous to the Resource Description Framework (RDF) format commonly used in the representation of knowledge in the Web Ontology Language (OWL). We choose this representation, as it will facilitate the comparison of claims across papers. To achieve this, we use the Python library \verb|textacy|, which is built on \verb|spaCy|~\cite{spaCy2020} and has a built-in extraction method that does not require the specification of relations in advance.

\subsubsection{LLM-based triple extraction}
As a second method for triple extraction, we consider an approach based on LLMs. The main advantage of such an approach is that LLMs can take into account context when extracting triples from a sentence. However, one has to be cautious, as LLMs can hallucinate and one can thus never be certain regarding the factuality of generations. In fact, LLMs have been reported to struggle with niche concepts~\cite{Wursch2023}. Therefore, prior to using LLMs it is critical to assess LLMs for hallucinations during triple extraction. Three state-of-the-art LLMs are considered: \texttt{Mistral-7B-Instruct-v0.2}, \texttt{Meta-Llama-3-8B-Instruct} and \newline \texttt{Starling-LM-7B-beta} ~\cite{jiang2023mistral, starling2023, llama3modelcard}. Our selection of candidate models follows directly from the deployment constraints of the pipeline. Because our aim is a triple-extraction method that is scalable, replicable, and runnable on modest hardware (16GB VRAM GPUs; Section~\ref{implementation}), we restrict attention to open-weight models in the 7–8B parameter range. Additionally, we exclude proprietary API-only systems, as they compromise replicability and are costly at the scale of hundreds of thousands of documents. We deliberately chose three instruction-tuned models that cover three distinct pretraining lineages, and two alignment paradigms: supervised instruction tuning with RLHF for the Llama and Mistral models, and reinforcement learning from AI feedback for Starling. Whereas the Starling and Llama models have a context length of 8192, the Mistral model has a context length of 32,000. 

We note that unlike the spaCy-based triple extraction, for LLM-based extraction we do not first extract claims from the text. The LLM is prompted to extract triples from the full input text, as the strength of LLMs is to take into account the full context. Therefore, we do not want to remove this context by isolating the claims.

\paragraph{Few-shot learning prompts and fine-tuning.}

Generative LLMs per se are not specifically pretrained or instruction-tuned for triple extraction. In order to improve their performance at this task, the two common approaches are few-shot learning prompts~\cite{LLMFewShotLearners2020OpenAI}, or fine-tuning~\cite{OpenAI2018GPT1}. 

Few-shot learning is generally considered as simpler to implement, given that it is equivalent to providing examples of desired text transformation as part of the user prompt. We formulated several of such example prompts based on the existing literature examples, and selected the best performing one. The resulting prompt can be found in Appendix Section \ref{fewshotapdx}

Fine-tuning LLMs is generally believed to be more effective, but requires significantly more computational power and a substantial amount of fine-tuning examples that scales with the model size. It is possible to mitigate both problems using parameter-efficient fine-tuning (PEFT), which reduces both the computational power requirements and offers better generalization with less data~\cite{falissard-etal-2023-improving}. Specifically, we use LoRA ~\cite{hu2021lora}, which freezes the majority of model parameters and fine-tunes only the scaling matrix of a singular value decomposition of a random matrix added to the model weights.

\subsection{Post-processing} \label{post_process}
Given that the end goal is to compare triples across papers, we normalize triple representation in order to facilitate the matching of nouns likely to designate technologies of interest. Specifically, we:

\begin{enumerate}
    \item Lowercase all words in the triple;
    \item Remove triples where either the subject or object contains more than 6 words;
    \item Remove stopwords from the triples based on the list included in \verb|NLTK|;
    \item Remove any non-text characters;
    \item Lemmatize verbs and nouns in the triple;
    \item Remove words containing less than 3 characters;
\end{enumerate}

Both cutoff values in 2. and 6. were chosen empirically, based on manual inspection of triple nouns.

\subsection{Filtering} \label{filtering}
As a final step, we filter the triples to identify the key domain terms on which the downstream analysis depends. Specifically, subjects and objects that designate technologies or technology-adjacent concepts specific to the field of interest, as opposed to rare, generic, or domain-agnostic vocabulary. We apply three complementary criteria, each removing a different class of non-domain term: terms too rare to denote an established technology (Frequency), terms too generic to be field-specific (Bookcorpus), and terms that are not discriminative of any particular field (Entropy).


\subsubsection*{Frequency}
First, we filter out triples with subjects or objects that are too rare. We do so by considering a \textit{control corpus} of all papers from the last quarter of 2023, enriched by a sample of 400 papers per month from the period January 2015 - September 2023. For every term $t$, we then define $f_{cc, t}$, which is the number of papers in the control corpus that contains the term at least 3 times per page, on average. If a triple has a subject or object of which $f_{cc, t} < 5$ for all terms $t$ in the subject or object, we remove this triple. This filtering step aims at removing nouns that are not used sufficiently frequently to correspond to a potential technology.

\subsubsection*{Bookcorpus}
Second, we use the general-purpose Gutenberg book corpus to detect triples that are generic and thus carry little information for technological monitoring. Let $f_{b, t}$ define the number of books in which term $t$ appears at least 3 times per page. Furthermore, we denote the number of papers in the control corpus as $N_p$ and the number of books in the book corpus as $N_b$. Then, each term $t$ obtains the following score $s_{t}$:

\begin{equation}
  s_{i} =
    \begin{cases}
      log(\frac{f_{cc, i}}{N_p}) - log(\frac{f_{b, i}}{N_b})& \text{if $f_{b, i} > 0$}\\
      \infty & \text{if $f_{b, i} = 0$}
    \end{cases}       
\end{equation}

The score of a term thus increases when the frequency of the term in the book corpus is lower. Simultaneously, the score increases when the frequency of the term in the control corpus is higher. We keep the triples of which at least one word from the subject and object is in the top 10\% of the term scores, with cutoff determined empirically.

\subsubsection*{Entropy}
Last, we filter terms according to how concentrated their usage is across arXiv categories. A term that appears throughout the taxonomy, such as "methodology" or "lab", carries little information about the field of a paper. In contrast, a term confined to a few categories is field-discriminative and thus more likely to denote a technology or technology-adjacent concept. We quantify this concentration with the \textit{categorical occurrence entropy} of a term: the entropy of its occurrence pattern over arXiv categories. Terms concentrated in few categories yield low entropy and are retained, whereas terms spread across many categories yield high entropy and are removed.


Formally, let us denote the set of all arXiv categories as $\mathcal{C}$, then the categorical occurrence entropy for term $t$ can be calculated as denoted below.

\begin{equation}
    H_t = \sum_{c \in \mathcal{C}}{P(t \in p | p \in c) \cdot log(\frac{1}{P(t \in p | p \in c)})}
\end{equation}

Here, we denote a paper as $p$ and the individual arXiv categories as $c$.

\subsection{Noun stapling}\label{nounstapling}
At this point in the pipeline, we have triples involving nouns that designate with high probability technologies and technology-adjacent terms, conformed to a somewhat standard representation. However, in order to perform an emerging technology analysis through technological convergence, we need to connect technology-designating nouns that indicate similar technologies despite naming variations. We refer to this as \textit{noun stapling}. 

When stapling (compound) nouns, we want to group subjects or objects together that are semantically sufficiently similar. As we are dealing with big datasets, we use a syntactic string similarity measure, which is fast and can combine detection of similarities at token and character level. Appendix \ref{app:semanticnounstapling} studies semantic embedding similarity as a measure for noun similarities, but we find that in niche domains embeddings are not able to capture similarities accurately, which is in line with the findings of \citet{Wursch2023}.

Specifically, we use a novel string similarity measure, introduced by \citet{Syntactic_string_measures}, that utilizes a combination of character- and token-level similarity measures. Specifically, the framework uses the \textbf{soft cardinality of a set}, as introduced by \citet{vargas2008knowledge}. The aim of a soft cardinality measure is to capture the number of unique concepts in a string. For example, the set \textit{(dog, dogs)} should have a lower soft cardinality than the set \textit{(dog, church)}. 

The results of \citet{vargas2008knowledge} showed that, in general, the combination of Dice (token-level) and 3-gram (character-level) worked well in all settings. Therefore, we choose to use this procedure, where we tokenize the compound nouns by splitting on spaces. Let us first introduce the 3-gram similarity measure between the strings $s_1$ and $s_2$, which is defined as 

\begin{align}
    QGramsSim(s_1, s_2) &= 1 - \notag\\[0.5em]
    &\hspace{-6.5em}\frac{\sum_{i=1}^{n}\big|match(q_i, Q_{s_1}) - match(q_i, Q_{s_2})\big|}{|Q_{s_1}| + |Q_{s_2}|},
\end{align}

where $Q_{s_1}$ and $Q_{s_2}$ are the sets of q-grams from $s_1$ and $s_2$ respectively. Furthermore, $n = |Q_{s_1} \cup Q_{s_2}|$, and $match(q_i,Q_{s_1})$ is the number of times
the q-gram $q_i$ appears in $Q_{s_1}$. This measure can then be used to calculate the soft cardinality of a set $T=(T^1, T^2, ..., T^n)$, which is defined as

\begin{align}
|T|_{soft} = \sum_{i=1}^{n}{\frac{1}{\sum_{j=1}^{n}{QGramsSim(T^i, T^j)}}}
\end{align}

Using the soft cardinalities, one can then compute the similarity between the two compound nouns $t_1$ and $t_2$ through an adjustment of the token-level measure. For the DICE measure, this then becomes:

\begin{align} \label{softdice}
SoftDICE(t_1, t_2) = \frac{2 \times |t_1 \cap t_2|_{soft}}{|t_1|_{soft} + |t_2|_{soft}}
\end{align}

Here, the soft cardinality of a set intersection can be computed as

\begin{align}
|t_1 \cap t_2|_{soft} = |t_1|_{soft} + |t_2|_{soft} - |t_1 \cup t_2|_{soft}
\end{align}
Last, one must choose a threshold above which the two compound nouns $t_1$ and $t_2$ are considered to be similar. As we want to be conservative, we choose a threshold of 0.85. Above this threshold, one can be relatively certain that two compound nouns are sufficiently similar to be marked as such.

\begin{table*}[pos=bp]
\caption{Pseudo algorithm for keyword clustering, based on the soft dice similarity as introduced in equation \ref{softdice}.}
\label{tab:pseudo_algo}
\begin{tabular}{@{}ll@{}}
\toprule
\multicolumn{2}{l}{\textbf{Pseudo algorithm:} keyword clustering based on similarity}                   \\ \midrule
1:  & \textbf{Inputs:}                                                                                  \\
2:  &   \quad \textit{keywordsList:} A list of keywords ordered by extraction count                             \\
3:  & \quad \textit{similarKeywords:} A dictionary which gives, for every keyword, a list of similar keywords \\
4:  & \textbf{Output:}                                                                                  \\
5:  & \quad \textit{Grouped keywords:} A dictionary of groups of keywords                                                                         \\
6:  & assignedKeywords $\leftarrow$ empty set                                                  \\
7:  & groups $\leftarrow$ empty dictionary                                                     \\
8:  & \textbf{for} each keyword1 in keywordsList \textbf{do}                                                     \\
9:  & \quad \textbf{if} keyword1 in groups \textbf{then}                                                               \\
10: & \qquad \textbf{continue}                                                                                 \\
11: & \quad \textbf{end if}                                                                                   \\
12: & \quad currentGroup $\leftarrow$ empty list                                                     \\
13: & \quad add keyword1 to currentGroup                                                             \\
14: & \quad \textbf{for} each keyword2 in similarKeywords{[}keyword1{]}                                       \\
15: & \qquad \textbf{if} keyword2 is in assignedKeywords \textbf{then}                                                  \\
16: & \qquad \quad \textbf{continue}                                                                                 \\
17: & \qquad \textbf{end if}                                                                                   \\
18: & \qquad add keyword2 to currentGroup                                                             \\
19: & \qquad add keyword2 to assignedKeywords                                                         \\
20: & \quad \textbf{end for}                                                                                  \\
21: & \quad groups{[}keyword1{]} $\leftarrow$ currentGroup                                           \\
22: & \quad add keyword1 to assignedKeywords                                                         \\
23: & \textbf{end for}                                                                                  \\
24: & \textbf{return} groups                                                                            \\ \bottomrule
\end{tabular}
\end{table*}

 \subsection{Key term extraction and clustering} \label{Key_term_extraction_methodology}
 From the triple extraction pipeline, we want to create a large graph of technology-related topics (nodes). To derive insights from this graph, it is essential to focus on a specific domain of interest. To this end, one requires a list of relevant topics. The manual curation of such a list requires a strong domain expertise, which is regularly not readily available. 
 
 Therefore, we develop a separate technology-designating keyword extraction for the first step of our approach. After manually specifying a few terms that broadly specify the field of interest, we select papers from arXiv that contain at least one of these terms in the abstract or title. Then, we extract a maximum of 10 key terms for each abstract using KeyBERT, with \texttt{allenai/specter} as the embedding model~\cite{grootendorst2020keybert, cohan2020specterdocumentlevelrepresentationlearning}.

 Next, we use the noun stapling method as described in Section \ref{nounstapling} to calculate similarities between the extracted key terms. We then use the pseudo-algorithm as given in Table \ref{tab:pseudo_algo} to cluster the key terms into \textbf{topics}. The algorithm takes a simple approach, leveraging the extraction count for each term. Starting from the most popular term, we iterate over the terms and check whether they are already assigned to a group. If not, we make a new group to which we add the term and all similar terms that are not yet assigned to a group. Last, the groups are named based on the most common substring across the key terms. If two substrings are equally common, we prefer to choose the longest one as it is likely more informative. 
 
Finally, we map the extracted triples onto these topic clusters to obtain the technology network used in the downstream analyses. Each subject or object is assigned to a topic if at least one key term in that cluster, where similarity is again measured with the SoftDICE similarity of Equation \ref{softdice}, using the threshold of 0.85 as discussed in Section \ref{nounstapling}. We note that it is possible for subjects and objects to be assigned to multiple topics simultaneously, if they are matched to keywords from multiple topics. This results in a graph where each node represents a topic, and the nodes are connected through the associated triples. Specifically, the predicates of the triples form undirected edges between the topics.


\begin{table*}[t]
\caption{Comparative benchmarking of triple extraction methods on the golden dataset. A ($\uparrow$) in the column header means that higher values are preferred, whereas a ($\downarrow$) means lower values are better.}
\label{tab:spacy_vs_llm}
\resizebox{\textwidth}{!}{%
\begin{tabular}{lccccc}
\hline
                                 & \begin{tabular}[c]{@{}c@{}}Avg. number \\ of triples ($\uparrow$)\end{tabular} & \begin{tabular}[c]{@{}c@{}}Perc. with\\ correct format ($\uparrow$)\end{tabular} & \begin{tabular}[c]{@{}c@{}}Perc. inconsistent\\ with Levenshtein dist. 2 ($\downarrow$)\end{tabular} & \begin{tabular}[c]{@{}c@{}}Perc. inconsistent\\ with Levenshtein dist. 3 ($\downarrow$)\end{tabular} & \begin{tabular}[c]{@{}c@{}}Avg. time / line\\ (s) ($\downarrow$)\end{tabular} \\ \hline
Annotated triples               & 8.1                                                                & 100\%                                                                    & 7.2\%                                                                                 & 2.6\%                                                                                 & -                                                                 \\
spaCy extraction                 & 3.9                                                                & 100\%                                                                    & 0\%                                                                                   & 0\%                                                                                   & 0.013                                                             \\
Starling - base                  & -                                                                  & 0\%                                                                      & -                                                                                     & -                                                                                     & 0.206                                                             \\
Mistral - base                   & -                                                                  & 0\%                                                                      & -                                                                                     & -                                                                                     & 0.221                                                             \\
Llama - base                     & 8.0                                                                & 10\%                                                                     & 0\%                                                                                   & 0\%                                                                                   & 0.097                                                             \\
Starling - base + few-shot       & 13.0                                                                & 100\%                                                                    & 13.2\%                                                                                & 4.6\%                                                                                 & 0.217                                                             \\
Mistral - base + few-shot        & 5.3                                                                & 60\%                                                                     & 28\%                                                                                  & 16\%                                                                                  & 0.243                                                             \\
Llama - base + few-shot          & 15.3                                                                & 100\%                                                                    & 7.9\%                                                                                 & 0.6\%                                                                                 & 0.104                                                             \\
Starling - fine-tuned + few-shot & 10.8                                                                & 100\%                                                                    & 40.8\%                                                                                & 24.8\%                                                                                & 0.220                                                             \\
Mistral - fine-tuned + few-shot  & 17.0                                                                & 75\%                                                                     & 50.4\%                                                                                & 41.1\%                                                                                & 0.244                                                             \\
Llama - fine-tuned + few-shot    & 13.4                                                                & 30\%                                                                     & 29.8\%                                                                                & 21.3\%                                                                                & 0.102                                                             \\ \hline
\end{tabular}%
}
\end{table*}

\subsection{Graph analyses}
Through the extraction and clustering of key terms in the domain of interest, and assigning the processed and filtered triples to the corresponding key topics, we have built a network of technological topics. We then perform several analyses on the network to derive useful insights from it. 

First, we identify established related technology clusters by partitioning the network into densely connected sub-networks through the Louvain community detection method~\cite{blondel2008LouvainMethod}. Specifically, we use the implementation in Gephi with a resolution of 0.85 \cite{gephi}. We scale the nodes based on the number of triples corresponding to the topics. The labels are scaled through the eigenvector centrality, which measures the influence a node has in a connected network. 

To assess technology convergences over time, we consider the topic co-publication frequency. Specifically, we use the Jaccard similarity between topics. Let $A(t)$ and $B(t)$ be the sets of triplets that are related to topics $A$ and $B$ at time $t$, respectively. Then, the Jaccard similarity 
 between the topics $A$ and $B$ at time $t$ is defined as 
\begin{align} \label{jaccard}
    J_{A, B}(t) = \frac{|A(t) \cap B(t)|}{|A(t) \cup B(t)|}
\end{align},
The Jaccard similarity will always be between 0 and 1, with a higher similarity indicating a high co-occurrence between the topics. A technology convergence will be identifiable by an increase in the Jaccard similarity.

\subsection{Implementation} \label{implementation}
To download the data, preprocess the text and extract the triples, we used 4 NVIDIA A100 Tensor Core GPU's with 40GB RAM. For the deployment of the pipeline, at least one GPU with 16GB of VRAM is required to run the LLMs. Depending on the size of the data, more GPUs may be required for swift processing. Furthermore, it is recommended to use at least 10 cores for swift downloading and preprocessing of arXiv papers and USPTO patent applications.

%% file: sections/results.tex
\begin{table*}[pos=b!]
\centering
\caption{Eight triples selected from the paper \textit{Simple Recurrence Improves Masked Language Models}~\cite{lei2022}, to illustrate the extraction, normalization and filtering of triples.}
\label{tab:triple-examples}
\small
\begin{tabular}{@{}p{0.30\linewidth} p{0.28\linewidth} p{0.34\linewidth}@{}}
\toprule
Raw triple (LM output) & Normalized triple & Outcome \\
\midrule
Transformer; rely; attention mechanisms
  & transformer; rely; attention mechanisms
  & \textbf{Retained} \\
Transformer architecture; interleave; multiheaded attention block
  & transformer architecture; interleave; multiheaded attention block
  & \textbf{Retained} \\
Transformer architecture; add; residual connection
  & transformer architecture; add; residual connection
  & \textbf{Retained} \\
scalar vectors; optimize; during training
  & scalar vectors; optimize; training
  & \textbf{Retained} (preposition \emph{during} dropped) \\
\midrule
works; demonstrate; superior performance
  & works; demonstrate; superior performance
  & \textbf{Removed} (entropy filtering of \emph{works}) \\
our model; perform; consistently
  & model; perform; consistently
  & \textbf{Removed} (entropy filtering of \emph{model}) \\
example; improve; translation
  & example; improve; translation
  & \textbf{Removed} (bookcorpus filtering of \emph{example} \\
model; achieve; average improvement
  & model; achieve; average improvement
  & \textbf{Removed} (entropy filtering of \emph{model} and \emph{average improvement}) \\
\bottomrule
\end{tabular}
\end{table*}

\section{Results} \label{Results}
Our results are organized into two distinct parts. In the first part (Section~\ref{extraction-method-result}) we compare the candidate triple-extraction methods and select the model that is used throughout the rest of the paper. The second part (Sections \ref{casestudy1} and \ref{casestudy2}) is applied: we deploy the selected pipeline end-to-end on two independent case studies: arXiv preprints (Section \ref{casestudy1}) and USPTO patent applications (Section \ref{casestudy2}). Here we show that it surfaces both established and emerging convergence patterns and that it generalizes across data sources.
\subsection{Selecting the Triple Extraction Method} \label{extraction-method-result}

In this first section, we will discuss the difference in performance between the spaCy and different LLM triple extraction methods, justifying our use of a specific LLM model and adaptation approach down the line. We are interested in four dimension of triple extraction performance: the number of triples extracted, the format of the generated text, signs of hallucination and the computation time.


 Although more extracted triples potentially mean a more granular technology identification, it is crucial to preserve their quality. Part of such assessment is evaluation of generative LLMs for hallucination instead of extraction, which we do by mapping the triples back to the reference text. Specifically, we consider whether the subjects and objects are present in the original text. We use the Levenshtein distance, which is defined as the number of single-character edits to change one word into the other. We note that the evaluation is performed on the raw triples produced by each extraction method, prior to the processing and filtering discussed in Sections \ref{post_process} and \ref{filtering}. This is essential, as the formatting and hallucination verifications require the raw output. The results, presented in Table \ref{tab:spacy_vs_llm} shows that there are also inconsistencies in the human-annotated triples. Through manual inspection, we identify that these inconsistencies in the human-annotated triples are caused by the stemming of verbs or nouns. This shows that the Levenshtein distance can cause false positives in the hallucination detection. Given that this is limited to 7\% for the human annotated triples, we accept this ground level of noise.

Table \ref{tab:spacy_vs_llm} also shows that the spaCy-based extraction method retrieves, on average, 3.9 triples per 15 lines. In contrast, on average, the LLMs extract up to 17 triples per text segment. 

A second aspect of extracted triple quality evaluation is formatting. In order to automatically parse the LLM output to use the triples for a downstream analysis, we need the formatting to be respected, which we observe to be only occurring for few-shot learning prompted models. Additionally, we observe that fine-tuning the LLM degrades the percentage of correctly formatted generations for Mistral and Llama. We expect that this is caused by the models being already extensively fine-tuned and being positioned on a Pareto frontier, with degeneration kicking in with further fine-tuning \cite{bai2022constitutional}.

Overall, the \texttt{Meta-Llama-3-8B-Instruct} model with few-shot prompting alone performs best, as it produces a large number of triples that are correctly formatted, while showing few to no signs of hallucinations. The main disadvantage of using a LLM for triple extraction is the computational overhead, as it is 10 times slower than the spaCy-based triple extraction. For this reason, we do not consider even larger LLMs. However, as a sufficiently large number of triples of high quality is essential for downstream performance, we choose to proceed with the Llama model.

\subsection{Case study part 1: technology forecasting on LLMs through arXiv e-prints} \label{casestudy1}

\subsubsection{Key term extraction and clustering} \label{keyterm_extraction}

From the 278,625 arXiv e-prints, we extract a total of 20,822,710 triples after post-processing and filtering, creating to our knowledge the largest unbiased graph of technology-related semantic triples. Table \ref{tab:triple-examples} shows 8 examples of extracted triples, highlighting which ones were retained and which ones were removed in the filtering stage. Next, following the approach described in Section \ref{Key_term_extraction_methodology}, we extract key terms from those papers related to LLMs. Here, we consider the same 278,625 papers from between 2018-2024, from the arXiv categories as specified in Table \ref{tab:arxiv_cats}. Specifically, we choose those papers with an abstract that contains at least one of the terms \textit{large language model}, \textit{large language models} or \textit{llm}. The extraction results in a total of 53,337 key terms from 4,182 papers. Next, the key terms are clustered into topics, following the approach that is described in Section \ref{Key_term_extraction_methodology}. The following analyses focus on the subjects and objects of the extracted triples, as we found them to be most informative. Additionally, Appendix \ref{app:predicate_analysis} provides an analysis of the triple predicates.


\subsubsection{Domain overview}



We start by leveraging the triples to provide a comprehensive overview of current research on large language models (LLMs). For that, we analyze the 20 most prominent topics in the field, presented in Table \ref{tab:keywords}, along with the five most frequently occurring keywords associated with each topic. According to that table, the most prevalent topics are \textit{LLM training} and \textit{Conversational AI}, highlighting the significant focus on optimizing model performance and enhancing human-computer interactions. 

Beyond these, several other key areas emerge, including \textit{language understanding}, \textit{question answering}, and \textit{retrieval-augmented generation (RAG)}. Research in \textit{language understanding} focuses on refining LLMs’ ability to process, interpret, and generate natural language with greater accuracy and contextual awareness. Question answering remains a critical application of LLMs, driving advancements in knowledge extraction, reasoning, and response generation across various domains. Additionally, RAG has gained traction as a method for improving factual consistency by integrating external knowledge retrieval with generative models, mitigating issues related to hallucination and enhancing the reliability of generated content.

\begin{figure}[h]
    \centering
    \includegraphics[width=1\linewidth]{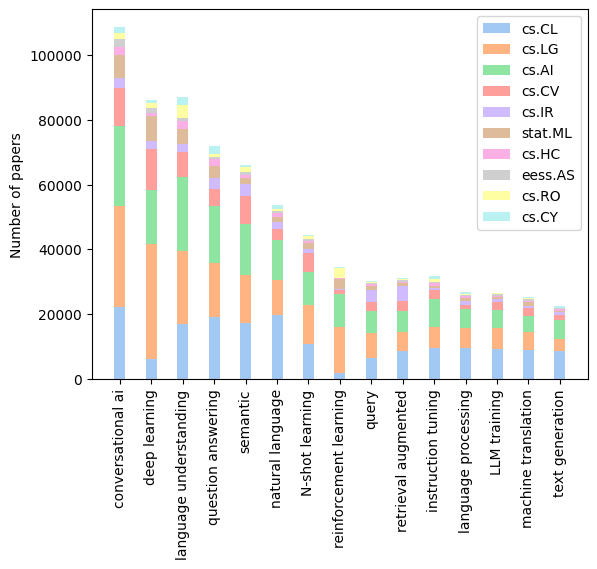}
    \caption{The number of papers for each of the 15 most common topics in the field of LLMs, based on the extracted and grouped key terms. Each of the bars is divided into the arXiv categories from which the papers originate.}
    \label{fig:stacked-barchart}
\end{figure}

To analyse the landscape in more detail, let us consider the size of the subtopics and their corresponding arXiv categories. Figure \ref{fig:stacked-barchart} shows the 15 largest topics and the corresponding arXiv categories from the papers to which the extracted triples belong. One can see that the most prominent arXiv categories are \verb|cs.LG| (machine learning) and \verb|cs.CL| (computation and language), but that there are also applications in \verb|eess.AS| (audio and speech processing) and \verb|cs.CR| (cryptography and security).

\begin{table*}[htbp]
    \caption{For each of the 20 largest topics in the field, the 5 most frequent key terms are displayed. The key terms are determined through the method described in Section \ref{Key_term_extraction_methodology}.}
    \centering
    \renewcommand{\arraystretch}{1.2} 
    \begin{tabular}{|p{4cm}|p{10cm}|}
        \hline
        \textbf{Category} & \textbf{Top 5 Keywords} \\
        \hline
        LLM training & trained language, training strat, training large, training large language, model pretraining \\
        \hline
        Conversational AI & generation, generative adversarial, conversation, generating, conversational \\
        \hline
        Deep Learning & deep learning, deep learning models, art deep, deep learning architectures, art deep learning \\
        \hline
        Language Understanding & understanding, planning, language understand, language understanding, language understandin \\
        \hline
        Question Answering & question, answering, question answer, question answering, question answerin \\
        \hline
        Semantic & semantic, semantic communication, automatic summarization, image semantic, semantic evaluation \\
        \hline
        Natural Language & natural language, natural language pro, natural language process, natural language processing, natural language inference \\
        \hline
        Retrieval Augmented & retrieval, retrieval augmented, retrieval augmented generation, generative retrieval, retrieval augmentation \\
        \hline
        Instruction Tuning & instruction, instructions, instruction following, instruction gen, based instruction \\
        \hline
        Language Processing & language processing, prompting large language, grounding language robotic, language modeling pretraining, prompted language \\
        \hline
        Machine Translation & translation, machine transla, machine translation, neural machine translation, translation model \\
        \hline
        Text Generation & generated text, al language generation, text generation, conversational agent, generation models \\
        \hline
        Video & videos, video generation, based video, video models, generated videos \\
        \hline
        Summarization & summarization, narrative, summarization model, summarization models, abstractive text \\
        \hline
        ChatGPT & chatgpt, based chatbot, chatgpt model, openai chatgpt, chatgpt1 \\
        \hline
        Multilingual & multilingual, multilingual bert, multilingual lan, multimodal large, multilingual large \\
        \hline
        Dialogue & dialogue, dialogue evaluation, dialogue task, dialogue tasks, dialogue modeling \\
        \hline
        Speech & text model, speech recognition models, text speech, speech text, based speech \\
        \hline
        N-Shot Learning & shot learning, shot learner, shot learners, zero shot, shot learning methods \\
        \hline
        Source Code & source code, code data, source code data, code test, source code sequence \\
        \hline
        Time Series & time series, time series forecasting, time series prediction, multivariate time series, time series dataset \\
        \hline
    \end{tabular}
    \label{tab:keywords}
\end{table*}

\begin{figure*}[h]
    \centering
    \includegraphics[width=\linewidth]{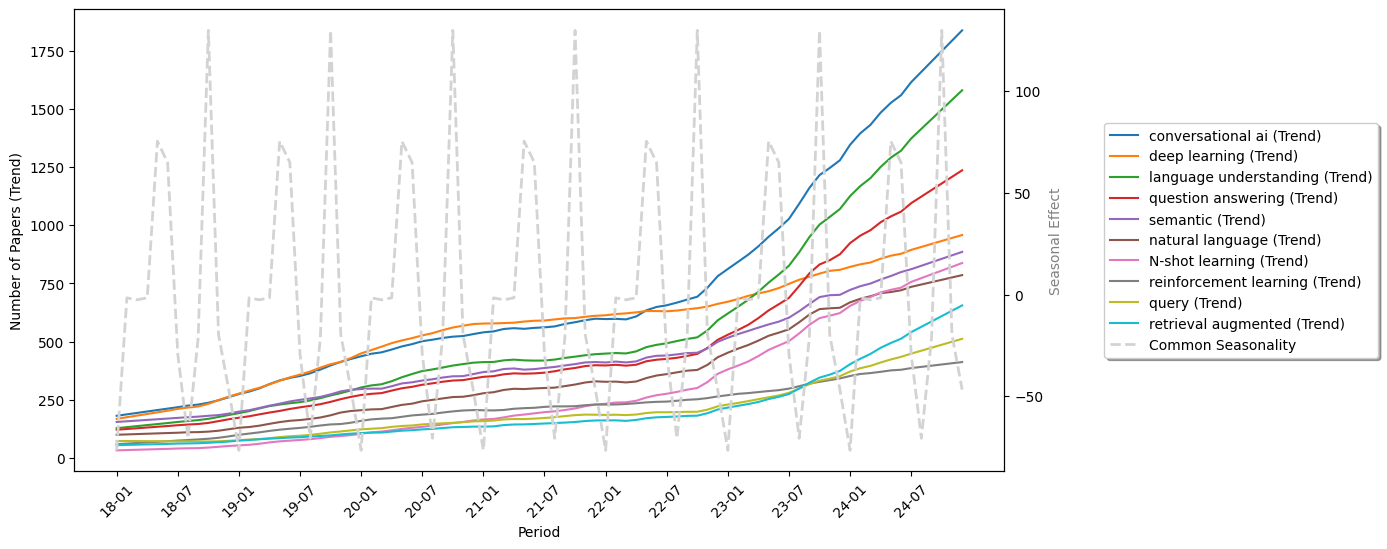}
    \caption{Multiple line plot showing for each of the 10 most common topics the number of papers in which at least one triple appears for that topic. The gray dashed line shows the aggregate seasonal component, whereas the colored lines show the trend for each topic. }
    \label{fig:tech_trend}
\end{figure*}

\subsubsection{Emerging technologies} \label{emerging_tech_arxiv}

Next, we consider emerging technologies by performing a time series analysis of the most popular topics. As publications follow yearly conference cycles, it is necessary to decompose the paper counts into a trend and a seasonal component. We therefore employ a simple seasonal decomposition based on moving averages. That is, we model the number of papers $N_p(t)$ at time $t$ for topic $p$ as

\begin{equation}
    N_p(t) = T_p(t) + S_p(t) + e_p(t),
\end{equation}

where $T_p(t)$ is the trend, $S_p(t)$ is the seasonal component and $e_p(t)$ is the residual. Figure \ref{fig:tech_trend} shows the trends for each of the ten most common topics, alongside the common seasonal component. We note that the trends are computed using separate seasonal components for each of the topics, but we display a common seasonal component for ease of interpretation. 

When considering the seasonal component, one can see two spikes a year, one around May-June and one in October. These spikes correspond to the biggest conferences for NLP related work: the Association for Computational Linguistics (ACL) and Empirical Methods in Natural Language Processing (EMNLP). The ACL conference is commonly around July, with submissions generally uploaded in the months before. The EMNLP conference is in November, with submissions uploaded to arXiv beforehand. 

When considering the trends of the various topics around LLMs on Figure \ref{fig:tech_trend} , one can see that, in general, there is a surge in interest for topics in LLMs in early 2023. This aligns with the release of ChatGPT in November 2022, which caused a surge in public interest in LLMs. Furthermore, this event seems to have had little effect in the popularity of older technologies such as \textit{deep learning} and \textit{reinforcement learning}. In contrast, the biggest emerging technologies are related to \textit{conversational ai} and \textit{language understanding}. This shows that the current focal point in the field is the simplified interaction between humans and LLMs.  

\begin{figure*}[H]
    \centering
    \includegraphics[width=1\linewidth]{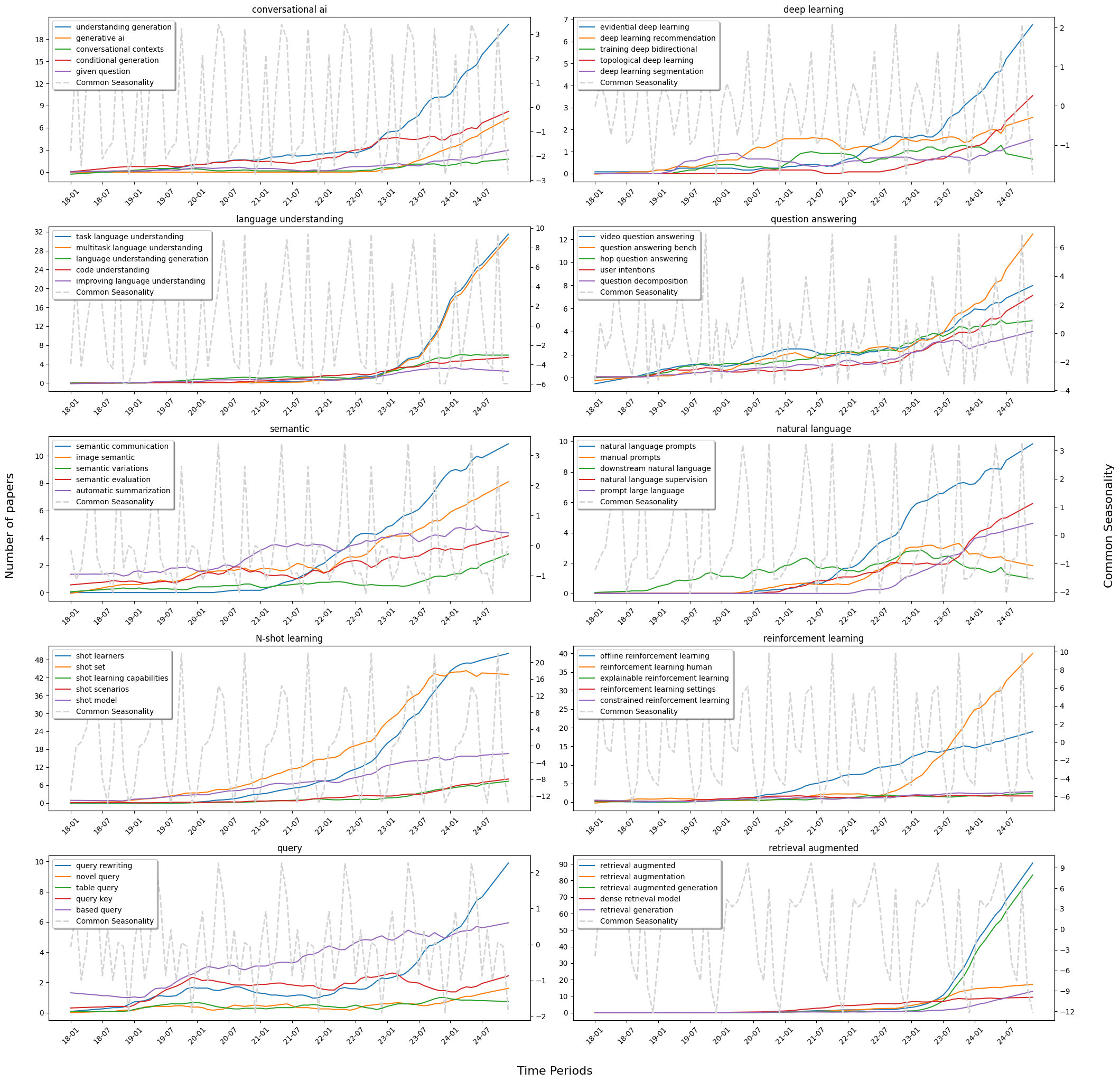}
    \caption{The trends for key sub-technology terms for the 10 most common emerging technology topics in the LLM space, based on the extracted and grouped key terms. The gray dashed line shows the aggregate seasonal component that the key terms share, whereas the colored lines show the trend for each key term.}
    \label{fig:tech_trend_grid}
\end{figure*}

For a more fine-grained analysis, we investigate the trends of individual key terms within the topics. Again, we distinguish between a seasonal component and the trend. To identify those key terms with the most interesting trends, we select the 5 key terms for each category that show the largest increase in prevalence between 2018 and 2024. Figure \ref{fig:tech_trend_grid} displays the trends for these key terms for each topic, alongside the aggregate seasonal components. The results show that there are specific research directions that surged over the past years, most notably \textit{evidential deep learning}, \textit{semantic communication}, \textit{reinforcement learning with human feedback} and \textit{query rewriting}, corresponding to emergent technologies. Such topics are all driven by advances in the capabilities of LLMs and are likely to continue growing in the near future.

\begin{figure*}[ht]
    \centering
    \begin{subfigure}[b]{0.45\textwidth}
        \centering
        \includegraphics[width=\linewidth]{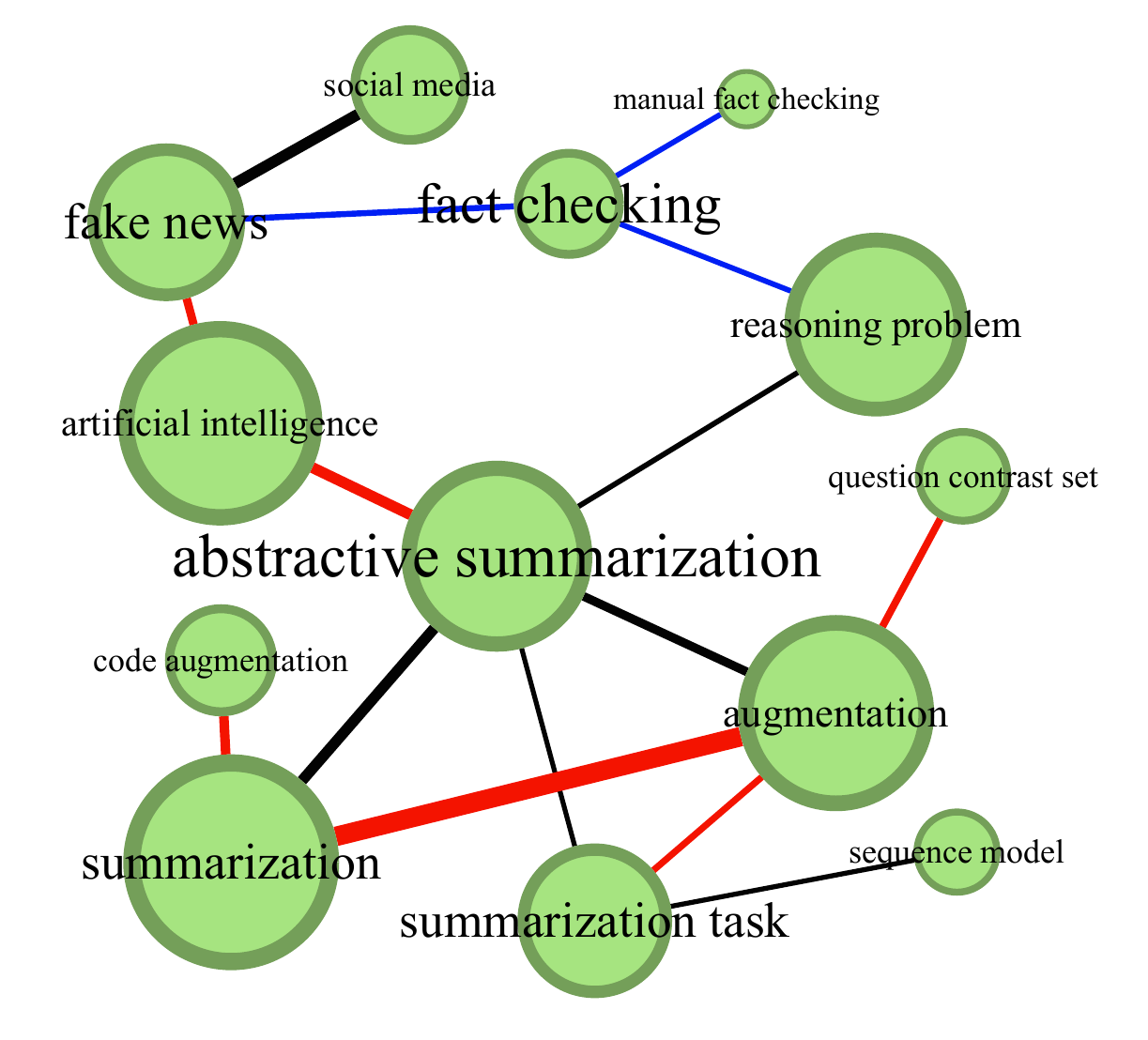}
        \caption{Summarization and factuality}
    \end{subfigure}
    \hfill
    \begin{subfigure}[b]{0.45\textwidth}
        \centering
        \includegraphics[width=\linewidth]{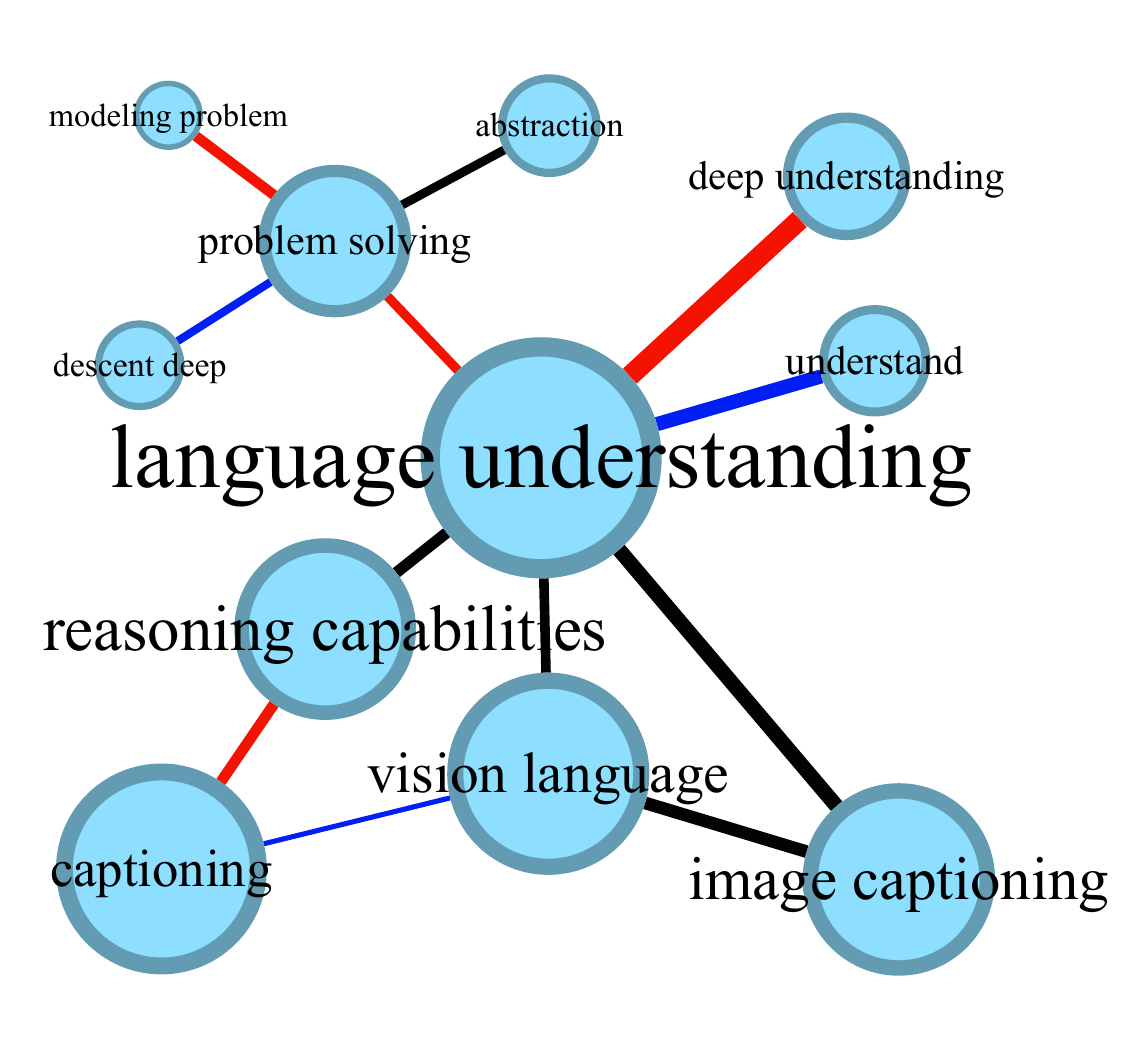}
        \caption{Language and image understanding}
    \end{subfigure}

    \vspace{0.5cm}

    \begin{subfigure}[b]{0.45\textwidth}
        \centering
        \includegraphics[width=\linewidth]{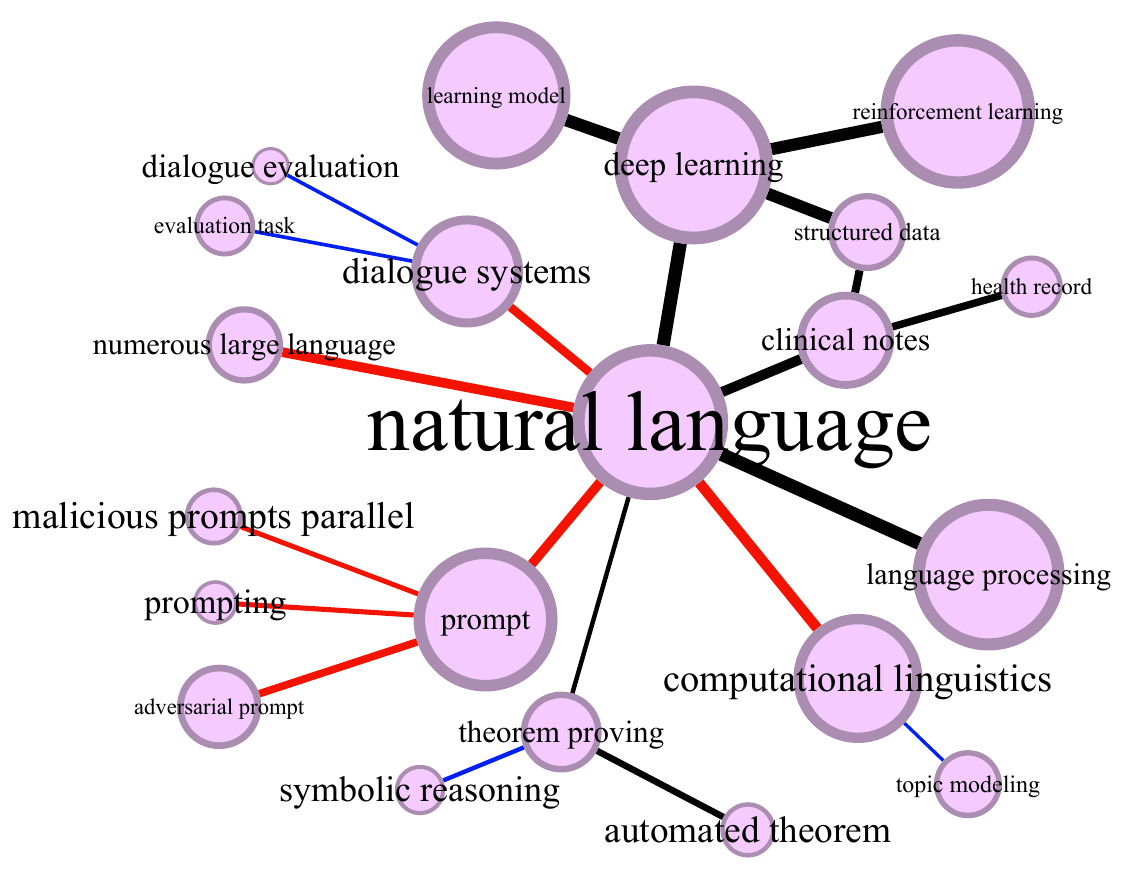}
        \caption{Core NLP concepts}
    \end{subfigure}
    \hfill
    \begin{subfigure}[b]{0.45\textwidth}
        \centering
        \includegraphics[width=\linewidth]{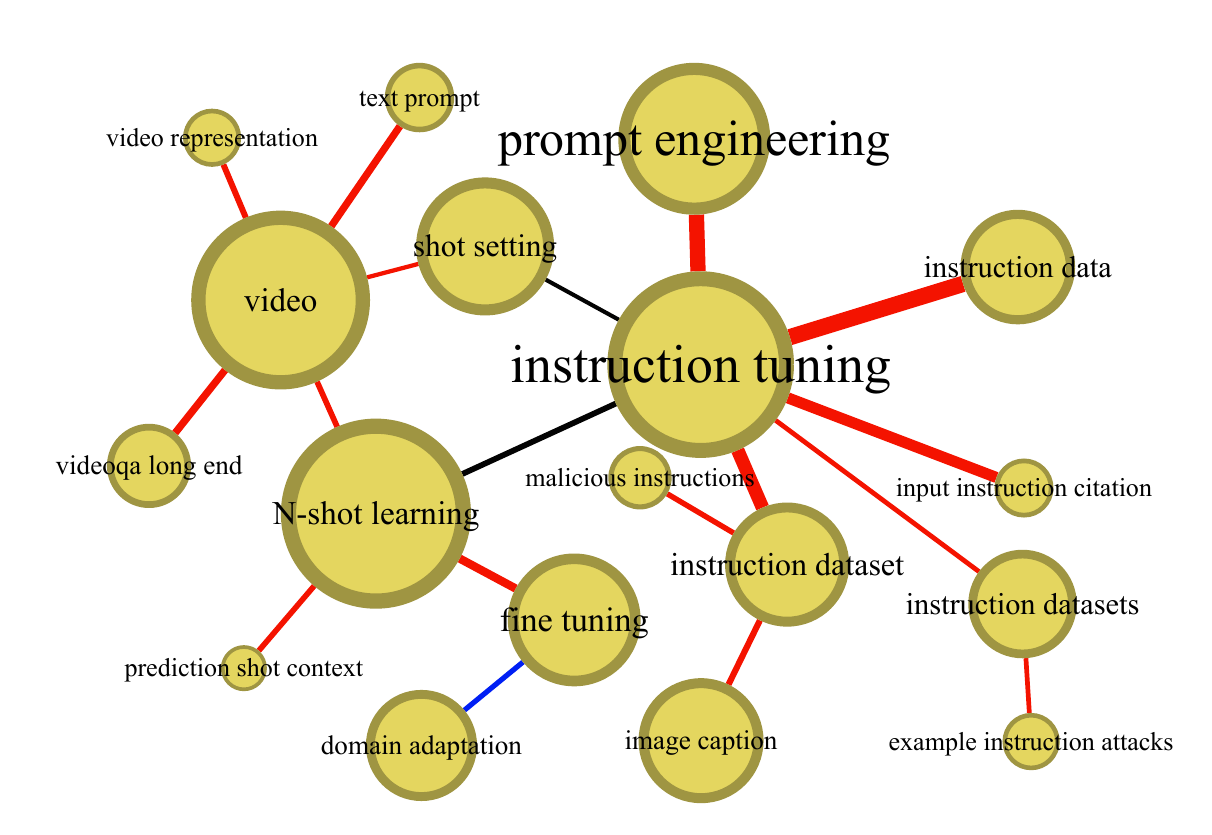}
        \caption{Instruction-tuning and prompting}
    \end{subfigure}

    \vspace{0.5cm}

    \begin{subfigure}[b]{0.45\textwidth}
        \centering
         \includegraphics[width=\linewidth]{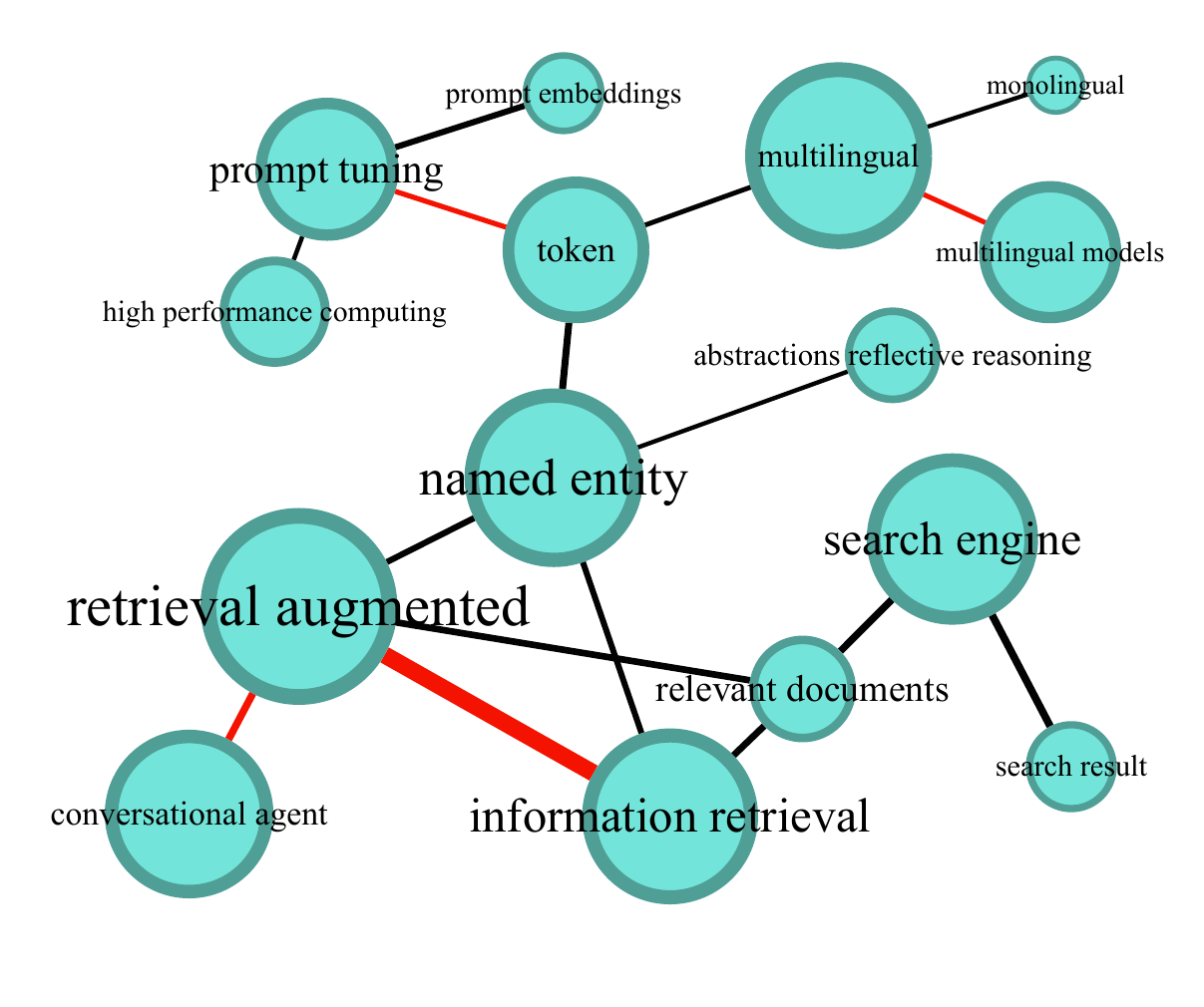}
        \caption{Retrieval augmentation}
    \end{subfigure}
    \hfill
    \begin{subfigure}[b]{0.45\textwidth}
        \centering
        \includegraphics[width=\linewidth]{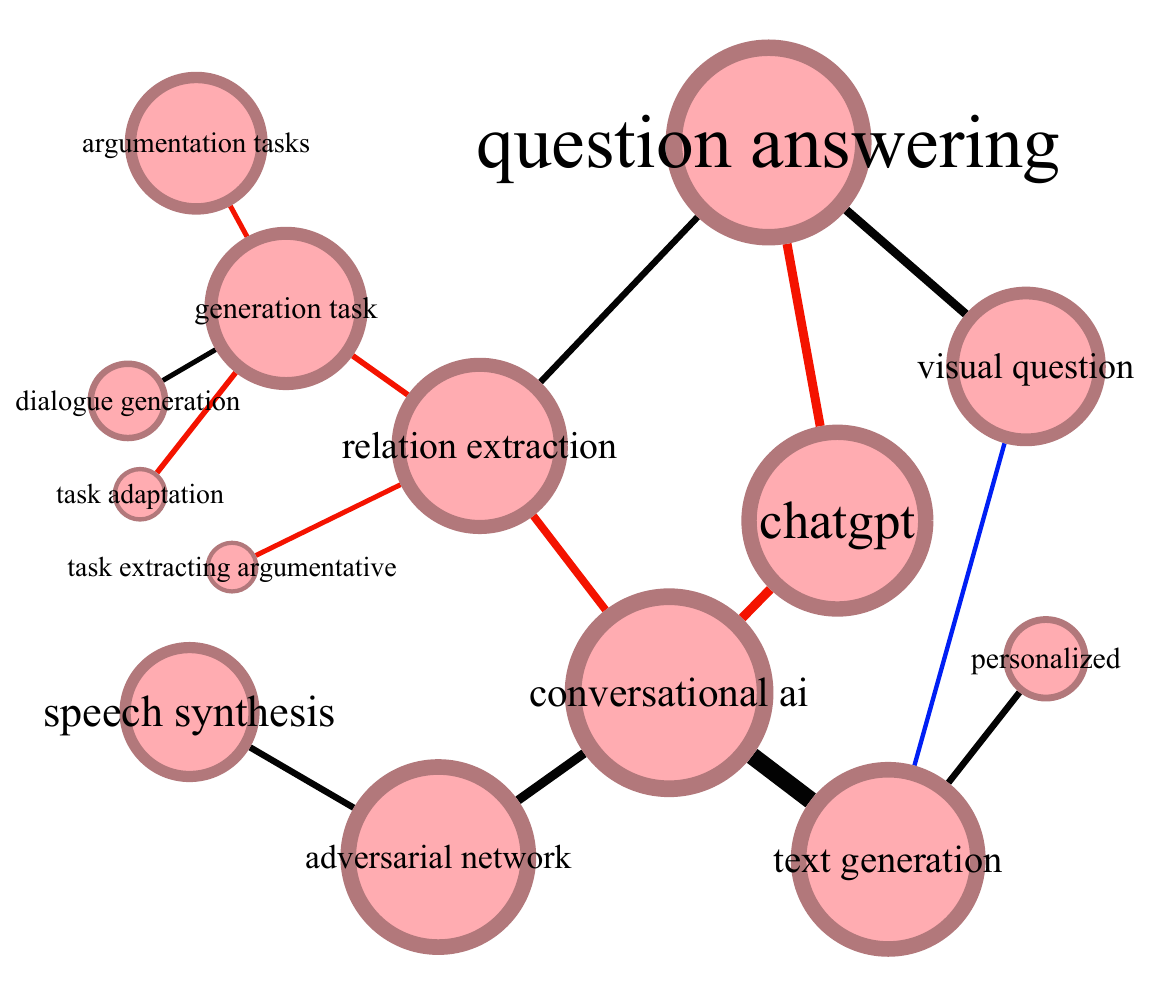}
        \caption{Question answering}
    \end{subfigure}

    \caption{Clusters of topics composed using the Louvain method for community detection. Relations in red occur over 70\% of the time in 2022 or later, whereas relations in blue occur over 70\% of the time in 2021 or earlier. All other relations are in black. Node/edge sizes correspond to absolute frequency, and text size to eigenvector network centrality.}
    \label{fig:clusters_network_arxiv}
\end{figure*}

\begin{figure*}[H]
    \centering

    \begin{subfigure}[b]{0.7\textwidth}
        \centering
        \includegraphics[width=\linewidth]{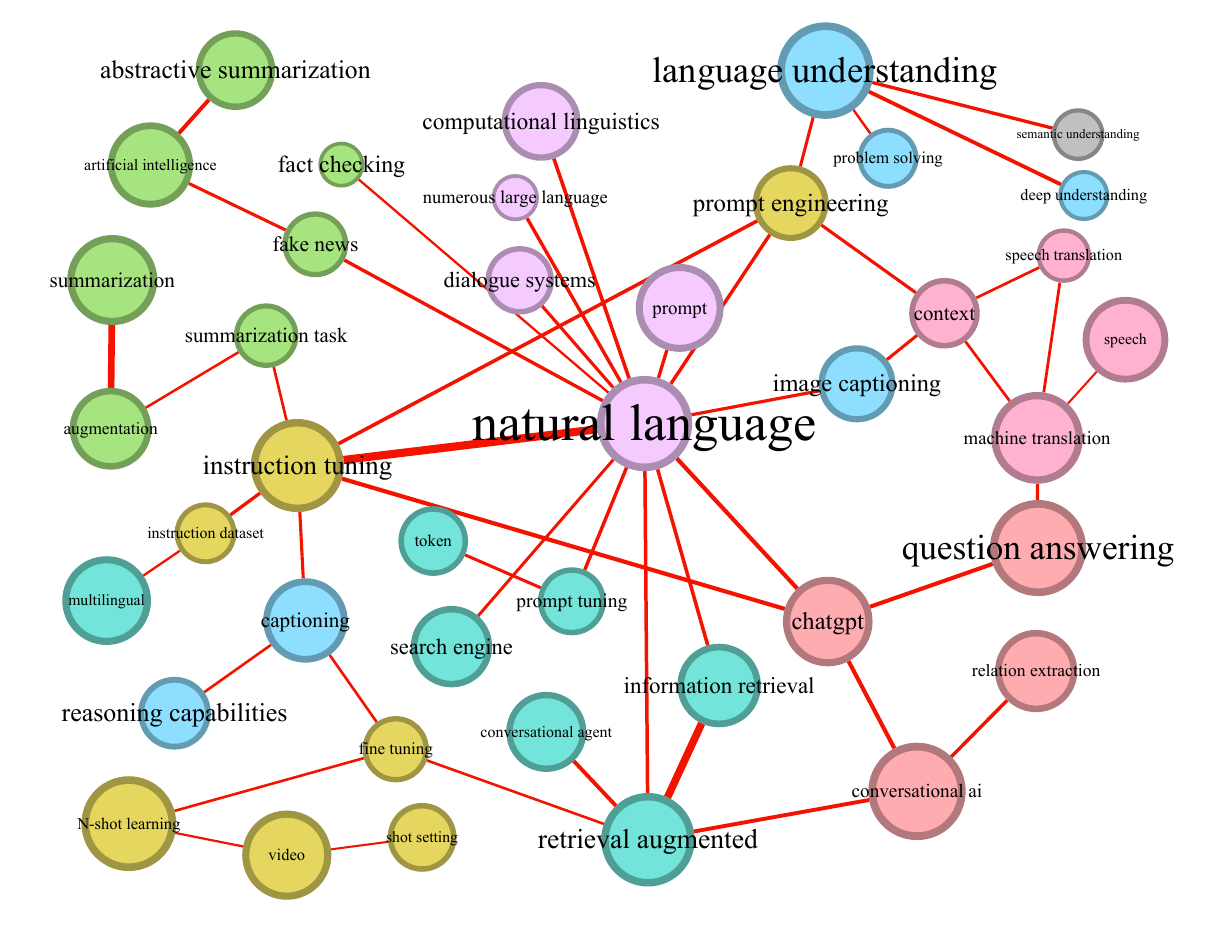}
        \caption{Emerging connections between key topics between 2018 and 2024.}
        \label{fig:emerging_synergies}
    \end{subfigure}
    \hfill
    \begin{subfigure}[b]{0.25\textwidth}
        \centering
        \includegraphics[width=\linewidth]{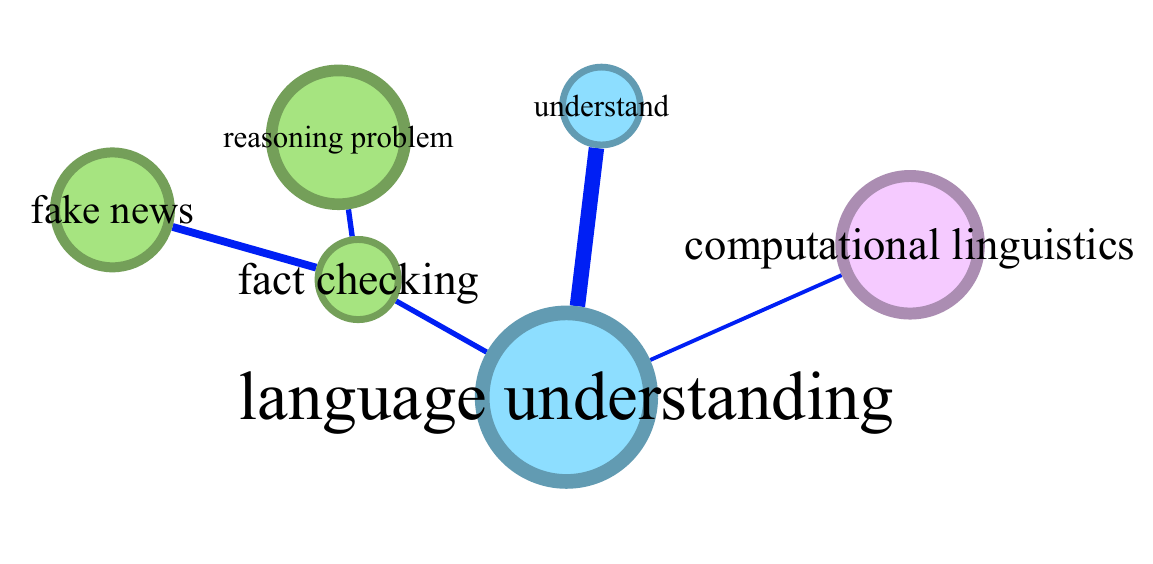}
        \caption{Connections between key topics that are losing prevalence between 2018 and 2024.}
        \label{fig:disappearing_synergies}
    \end{subfigure}

    \vspace{1em}

    \begin{subfigure}[b]{0.8\textwidth}
        \centering
        \includegraphics[width=\linewidth]{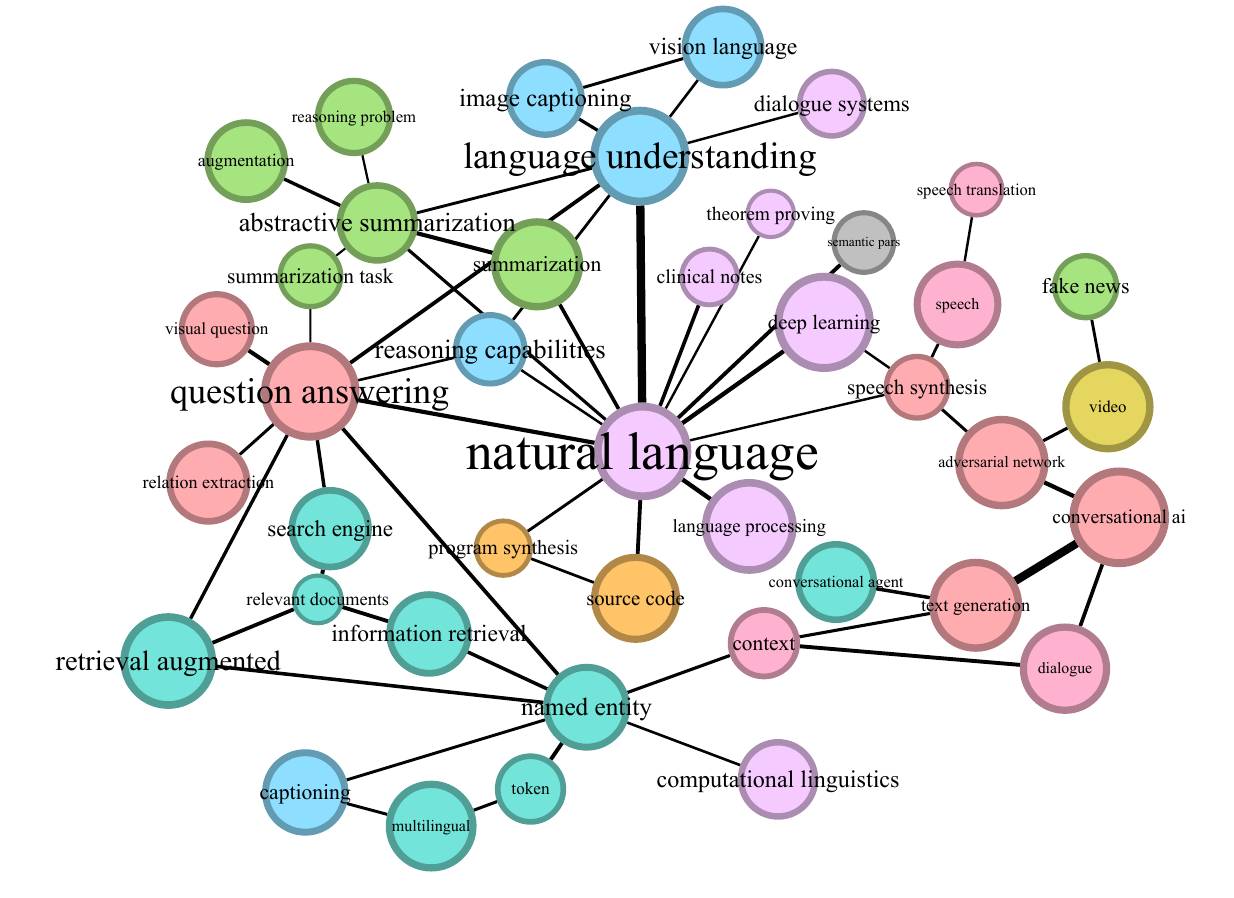}
        \caption{Persistent connections between key topics between 2018 and 2024.}
        \label{fig:persisting_synergies}
    \end{subfigure}

    \caption{Emerging (red), disappearing (blue), and persistent (black) relations between technology-related term clusters based on the arXiv paper data. Relations in red occur over 70\% of the time in 2022 or later, whereas relations in blue occur over 70\% of the time in 2021 or earlier. All other relations are displayed as black. The node color reflects the term cluster they belong to, the sizes of the nodes and edges reflect the frequency of their appearances, and the size of the text labels reflects the eigenvector network centrality.}
    \label{fig:trends_network_arxiv}
\end{figure*}

\subsubsection{Technology Convergence analysis}\label{sec:network}

The technology convergence analysis is the heart of our approach, and we perform it through topological analysis of semantic triple graph linking together technology-designating terms. We perform a network analysis, where each node represents a key topics and the edges represent the number of triplets connecting the topics, with edges mapping both to the connecting verb and the paper from which the triple was extracted. The edges are then aggregated, leading to a single weighted undirected edge, with weight proportional to triple occurrence. We also assign a "color" to the edge, based on whether the occurrence of triples involving the two technologies has been increasing (red), decreasing (blue) or remained consistent (black).

\paragraph{Technology clusters}

We identified established related technology clusters by partitioning the network into densely connected sub-networks, by using the Louvain community detection method~\cite{blondel2008LouvainMethod}\footnote{For legibility reasons we are unable to provide the entire network}. Figure \ref{fig:clusters_network_arxiv} shows the six largest detected clusters, with node color indicating the cluster. The size of nodes corresponds to the frequency of technology-related term occurrence, while the size of the term itself - to the node eigenvector centrality on the entire network. The eigenvector centrality corresponds how important the node is to connecting the entire network.



\begin{figure*}[H]
    \centering
    \includegraphics[width=1\linewidth]{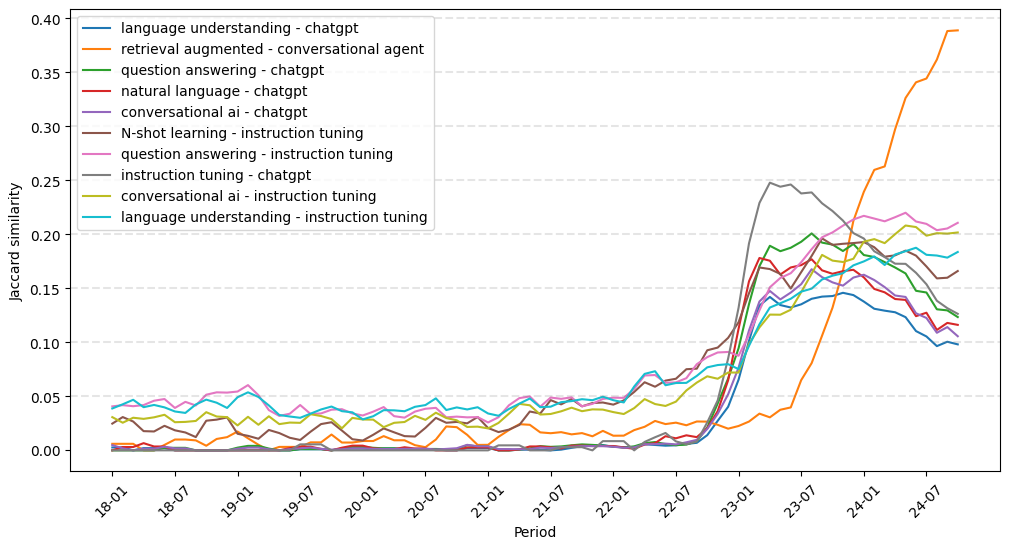}
    \caption{The Jaccard similarities for the 10 pairs of topics for which the Jaccard similarity had the largest increase between in the Jaccard similarity between 2018 and 2024. We consider all topics, as identified by the extracted and grouped key terms in for the topic of LLMs.}
    \label{fig:jaccard}
\end{figure*}

\paragraph{Technology convergences}

In order to identify emerging technological convergences, we then focus on the technology-related terms connecting different technology clusters. Figure \ref{fig:trends_network_arxiv} shows three networks, the first presenting emerging technological connections, the second showing waning ones, and the last one showing persisting ones.
We observe an emerging technological convergence between \textit{instruction tuning} and \textit{natural language}. This highlights the surge in conversational AI that we discussed previously, which was unlocked by instruction-tuned LLMs, that allowed general public access to conversational LLMs and made them useful as a human assistant emulators~\cite{ouyang2022training}. Furthermore, we see pivotal roles for \textit{prompt engineering} and \textit{retrieval augmentation}. These relatively new branches of research aim at improving the performance of LLMs. For instance, we see that retrieval augmentation is linked to information retrieval, illustrating its goal of improving the output of LLMs by injecting data from an existing database.

When we consider the relations that declined in prevalence between 2018 and 2024, we most notably see a link between \textit{language understanding} and \textit{fact checking}. This seems to suggest that fact checking is increasingly less related to language understanding. One hypothesis is that fact checking is now more often performed through stylistic patterns, leveraging ML methods.

Finally, we consider persistent relations between 2018 and 2024, which often involve core concepts in LLMs. For instance, \textit{question answering} links with \textit{language understanding} and \textit{reasoning capabilities}. Similarly, \textit{retrieval augmentation}, introduced early, has remained stable in its connection to \textit{question answering}.


\paragraph{Temporal trends of convergence}
Last, we analyze the temporal aspect of convergence by considering the topic co-publication frequency. To analyse which technologies are converging, we study the pairs of topics that have the largest increase in the Jaccard similarity between 2018 and 2024. The Jaccard similarity measure is defined in equation \ref{jaccard}. Figure \ref{fig:jaccard} shows the evolution of the Jaccard similarities for these pairs of topics. One can see that from the release of ChatGPT in November 2022, several technology convergences emerge rapidly. However, it can also be observed that in 2024 several of such convergences start to decline again, indicating that the effect was only temporary. In contrast, one can see that the convergence between \textit{retrieval augmented} and \textit{conversational agent} is stronger and is not yet stagnating. In combination with the network analysis that highlighted these two topics as technological convergences as well, we can designate \textit{retrieval augmentation} and \textit{conversational agents} as emerging transformative technologies in the field of LLMs, as of end 2024, based on research preprints published on arXiv.

\subsection{Case study part 2: technology forecasting on LLMs through USPTO patent applications}\label{casestudy2}

\begin{figure*}[H]
    \centering
    \includegraphics[width=1\linewidth]{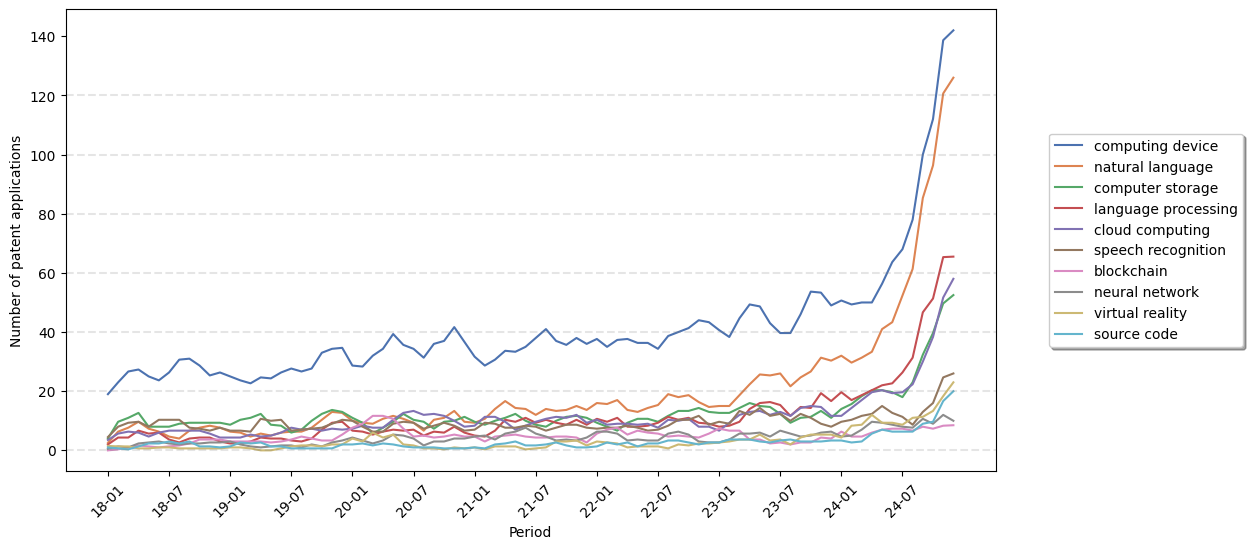}
    \caption{Multiple line plot showing for each of the 10 most common topics, as identified by the extracted and grouped key terms, the number of patent applications in which at least one triple appears for that topic. }
    \label{fig:patent_line_plot}
\end{figure*}

In order to demonstrate the generalization of our approach and its usefulness in the fields with less emphasis on early preprint publication, we use USPTO patent data to provide a different perspective on the technological developments related to LLMs. From the 9,793 patent applications, our pipeline extracts 3,027,121 triples after post-processing and filtering. As before, to perform an informative downstream analysis, we require a set of key terms that are most relevant to the field of LLMs. For this purpose, we extract key terms from the abstracts of the 9,793 patents, following the same methodology as described in Section \ref{Key_term_extraction_methodology}.

We focus on two analyses to showcase the unique perspective that patents give on the technological developments surrounding LLMs compared to arXiv preprints. As the primary goal of this section is to highlight the generalizability of our method across different data sources, we do not cover all of the perspectives that were taken using the arXiv papers, but merely highlight two approaches.

\subsubsection{Emerging technologies}
We start by performing a technology-related term usage  analysis similar to Section \ref{emerging_tech_arxiv}, where we are looking for terms with recently increasing usage. Whereas papers uploaded on arXiv follow a strong seasonal pattern due to yearly conference cycles, this pattern is not present for patent applications, due to the lack of yearly conference deadlines.

Figure \ref{fig:patent_line_plot} shows the most popular emerging terms associated with LLMs in patent applications. Across the topics, there is a surge in LLM-related technology terms halfway through 2024. This rise in frequency of term usage tracks the rise of similar terms usage on arXiv at the end of 2022, suggesting the analysis of scientific articles preprints allows an early trend emergence identification.

We see that LLMs are connected to a variety of fields, for instance \textit{computer storage}, \textit{cloud computing}, \textit{source code}, and \textit{speech recognition}. Furthermore, we note that the rise in patent applications has only started recently, and it will be interesting to see if future trends follow the patterns identified in the temporal analysis of arXiv preprints, with some rapidly growing technolgy topics pleateauing shortly. 

\begin{figure*}[!b]
    \centering
    \includegraphics[width=0.6\linewidth]{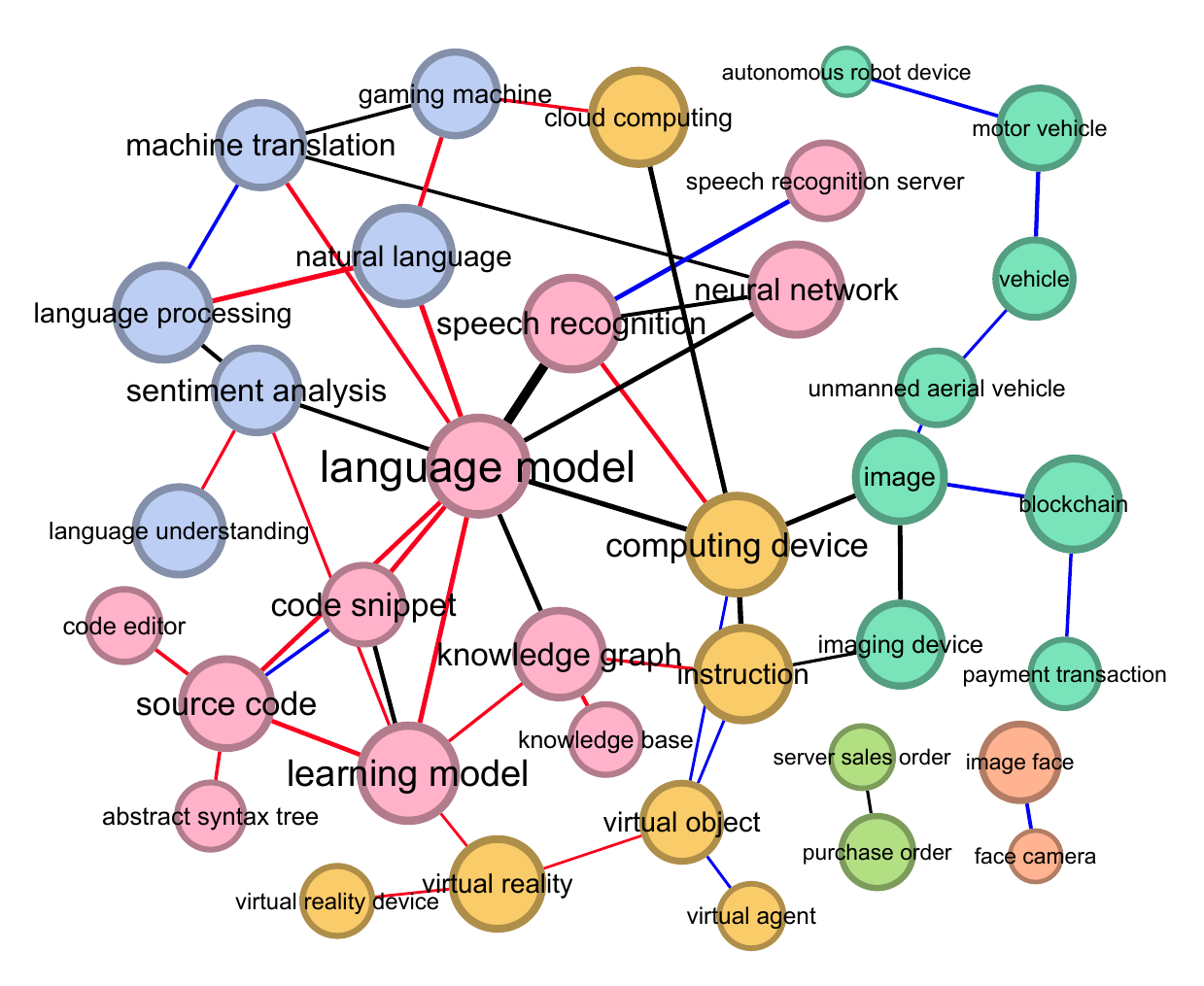}
    \caption{Network representation of the key topics based on the USPTO patent application data. Relations in red are occurring over 70\% of the time in 2022 or later, whereas relations in blue occur over 70\% of the time in 2021 or earlier. All other relations are displayed as black. The node color corresponds to the technology term cluster they belong to, the sizes of the nodes and edges correspond to the absolute frequency of their appearances, and the size of the text labels corresponds to the eigenvector network centrality.}
    \label{fig:patent_network}
\end{figure*}

\subsubsection{Network analysis}
Just as for arXiv articles, we analyze technological convergence through technology-related terms semantic graph analysis. Figure \ref{fig:patent_network} displays the network, with node and edge color and size corresponding to the same properties as for arXiv semantinc graph analysis in section~\ref{sec:network}.

From this visualization, we can make several observations regarding the developments in industry applications of LLMs. First, we see that there is a significant overlap between LLMs and self-attention models, covering multimodal models, visual attention models and speech attention models. For waning topic connections, we see that there is a decline in image processing connection to blockchain and payment processing, likely corresponding to the deflation of the non-fungible tokens (NFT) bubble; as well as to self-driving vehicle and unmanned aerial vehicles (UAVs), likely corresponding to a switch to other types of sensors and manual controls. The rupture of the link between speech recognition and speech recognition server seems to indicate the move of speech recognition onto devices.

When considering emerging topics, one can observe that there is a rise in interest for knowledge graphs and knowledge bases. Specifically, this seems to be closely related with the instruction-tuning of LLMs. Furthermore, one can see a rise in attention to source code manipulation, bridging the classical abstract syntax tree analysis with LLM-based machine learning methods, corresponding to the trend of LLM-based code generation and analysis integration into programming workflows. 

Based on the USPTO patent applications, our analysis suggests that \textit{coding LLMs} and \textit{on-device speech recognition} with self-attention models are emerging transformative topics in the field of LLMs and self-attention models, as of end 2024.

%% file: sections/conclusion.tex
\section{Conclusion} \label{Conclusion}

The monitoring and forecasting of emerging transformative technologies is a notoriously important and yet difficult topic in the field of technological forecasting. The accelerated accumulation and availability of potentially relevant data presents an opportunity for more grounded forecasts, but also poses a challenge due to the scale and presence of noise. We developed and presented in this paper a pipeline that leverages advances in NLP and big data analysis to perform a technology convergence analysis while addressing the challenges that previously made such analysis at scale impossible. Specifically, we (I) leveraged the recent advances in NLP and recently published benchmarking datasets for to extract information relevant to technological forecasting in an unbiased structured format, and at a scale that has been previously unattainable; (II) developed a new approach to account for variation in technologies designation through varying terms allowing their systematic analysis; (III) constructed a novel graph-based technique for identification of emerging technologies and likely transformative emerging technologies for specific fields.

We then use the pipeline we developed in order to analyze the trends in the LLM-related technology field based on two different unstructured data sources on a previously unseen scale and granularity. Specifically, we perform a full-text analysis of 278,625 arXiv scientific preprints and 9,793 USPTO patent applications, resulting in 53,337 LLM technology-related terms and 23,849,831 triples. By leveraging big data techniques on the resulting semantic graph analysis, we were able to identify two emerging transformative technologies as of 2024: \textit{retrieval-augmented generation} and \textit{conversational agents}, which seem to be supported by the data-grounded agentic AI interest in 2025; as well as two emerging transformative applications: \textit{source-code generation and processing} with LLMs and \textit{on-device speech processing with multimodal models}. 

We leave to future research the task of validating the identified transformative technologies, assessing the pipeline’s use with other data sources — such as social media, blogs, and grey literature like reports and white papers — and examining its longer-term predictive capabilities.


%% file: sections/appendix.tex
\newpage
\onecolumn
\section{Appendix}

\subsection{Few-shot prompting} \label{fewshotapdx}
We used few-shot prompting for the triplet extraction. We attempted various different prompts and achieved the best results with the setting that is illustrated below. The \texttt{\textit{\textbf{input text}}} is the text for which we will extract the triples. 

\vspace{3mm}

\noindent\fbox{%
    \parbox{\textwidth}{%
    \begin{center}
    \textcolor{red}{\texttt{\textbf{User: }}} \texttt{You will extract the subject-predicate-object triples from the text and return them in the form [(subject\_1; predicate\_1; object\_1), (subject\_2; predicate\_2; object\_2), ..., (subject\_n; predicate\_n; object\_n)]. I want the subjects and objects in the triples to be specific, they cannot be pronouns or generic nouns. The first text is: Ever since the Turing Test was proposed in the 1950s, humans have explored the mastering of language intelligence by machine. Language is essentially a complex, intricate system of human expressions governed by grammatical rules.} \\
    \textcolor{blue}{\texttt{\textbf{Example output:} }} \texttt{[(turing test; proposed in; 1950s), (human; explored; language intelligence), (language; is; system of human expressions), (grammatic rule; govern; language)]} \\
    \textcolor{red}{\texttt{\textbf{User: }}}\texttt{The next text is: Currently, LLMs are mainly built upon the Transformer architecture, where multi-head attention layers are stacked in a very deep neural network. Existing LLMs adopt similar Transformer architectures and pre-training objectives (e.g., lan-guage modeling) as small language models.} \\
   \textcolor{blue}{\texttt{\textbf{Example output:} }} \texttt{[(large language model; built upon; transformer architecture), (multi-head attention layer; stacked in; deep neural network), (large language model; adopt; transformer architecture)]} \\
    \textcolor{red}{\texttt{\textbf{User: }}}\texttt{The next text is: After collecting a large amount of text data, it is essential to preprocess the data for constructing the pre-training corpus, especially removing noisy, redundant, irrelevant, and potentially toxic data, which may largely affect the capacity and performance of LLMs.}\\
   \textcolor{blue}{\texttt{\textbf{Example output:} }}  \texttt{[(preprocessing; remove; noisy data), (preprocessing; remove; redundant data), (preprocessing; remove; irrelevant data), (preprocessing; remove; toxic data), (noisy data; affect; performance large language model), (redundant data; affect; performance large language model), (irrelevant data; affect; performance large language model), (toxic data; affect; performance large language model)]} \\
    \textcolor{red}{\texttt{\textbf{User: }}}\texttt{The next text is:} \textbf{\textit{input text}}\\
    \textcolor{green}{\texttt{\textbf{Assistant:} }}  \textbf{\textit{Output}} \\
    \end{center}
}}

\subsection{Threshold Selection} \label{app:thresholdselection}
Here, we provide justification for several threshold selections for filtering and noun stapling, as described in Sections \ref{filtering} and \ref{nounstapling}. Specifically, we discuss the length threshold, frequency threshold, and the SoftDICE threshold. 

\subsubsection{Post-processing of Triples}\label{app:thresholdselection_postprocessing}
\paragraph{Subject and Object length.} We first discuss the choice for removing subjects and objects with more than 6 terms. We opted for this threshold through manual inspection of the samples. Specifically, we considered 50 randomly selected entities of lengths 1 through 10. Here, in Table~\ref{tab:samples-length} we show a representative sample of 10 entries from length 4 to 8. This shows that above length 6, entries are never clear entities, and are safe to remove without losing information. We note that for length 5 and 4, there are also entries that are no clear entities. However, we choose to be conservative with the threshold choice to not lose information, and also rely on subsequent filtering steps. 

\begin{figure}
    \centering
    \includegraphics[width=0.5\linewidth]{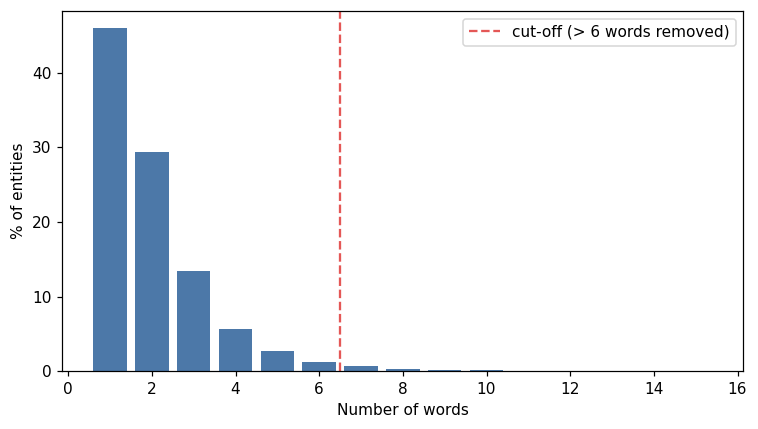}
    \caption{Distribution of the subjects and objects versus the number of words they contain.}
    \label{fig:frequency}
\end{figure}

Additionally, Figure \ref{fig:frequency} shows the percentage of entities with a specific number of terms. We see that most entities have 1 or 2 terms, and we are removing only a small fraction of subjects and objects in this filtering step. This is also expected, as the few-shot examples should induce the LLM to not often extract subjects and objects with multiple terms.

\begin{table}[h]
\centering
\caption{Representative subjects/objects by length. At lengths 4-6 the spans are
still mostly (compound) technology entities alongside some noun phrases; at lengths 7-8 they tend to be clauses that no longer name a single entity.}
\label{tab:samples-length}
\small
\begin{tabular}{cp{0.85\linewidth}}
\toprule
Length & Examples \\
\midrule
4 & deep feedforward neural network; neural machine translation systems; multi-agent markov decision process; bayesian experimental design algorithm; instance-wise adaptive adversarial training; user interest-aware capsule network; machine learning pipeline management; renewable energy sources generation; concentration of particulate matter; number of oracle calls \\
\addlinespace
5 & fully connected deep neural network; semi-markov conditional random fields model; exponential family variational kalman filter; hierarchical attention multi-task learning networks; improved phishing spam detection model; other graph contrastive learning methods; balance between efficiency and quality; discrepancy between training and inference; finite amount of training data; model performance and training time \\
\addlinespace
6 & security information and event management system; gumbel tree-based long short-term memory network; bounded information rate variational autoencoder algorithm; high resolution image synthesis and generation; computer vision and natural language processing; signal processing and machine learning techniques; competitiveness and effectiveness of proposed method; powerful representation capability of neural networks; detecting misinformation at an early stage; analysis on impact of network parameters \\
\addlinespace
7 & models may be sensitive to few epochs; weights do not move far from initialization; probability of document belonging to each class; disparities extend far beyond our evaluated domains; reason the intended purpose of the code; rewards linked to amount of pixels acquired; distance from data points to decision boundary; effect of number of sampled image patches; quick adaptation to patterns in input prompts; model more sensitive to relative positional information \\
\addlinespace
8 & more training data helps to improve certified accuracy; repairs do not impact large language model performance; approach performs well on synthetic and real-world datasets; preserving direction of momentum leads to better performance; using convolutional neural networks to solve poisson equation; whether one object is completely inside the other; how many countries are diplomatically related to italy; required for emergency vehicle to arrive at destination; power of each negotiator building on its history; importance of aligning storytelling approach with customer journey \\
\addlinespace
\bottomrule
\end{tabular}
\end{table}

\paragraph{Word length.}
Second we discuss the choice to remove words with 2 characters or less. Table \ref{tab:character_count} shows 10 randomly sampled 1-character, 2-character and 3-character terms from the raw triples from the arXiv papers. We see that the 1-character and 2-character terms tend to be noise or stopwords, such as \textit{of} and \textit{in}. For the 3-character terms, there are also filler terms (such as \textit{and} and \textit{the}). However, there are also meaningful terms included, such as \textit{fit}, \textit{sum} and \textit{low}. Therefore, we choose to be conservative and choose to remove terms with 2 characters or less. We note that due to incomplete abbreviation resolution, there may still be two character abbreviations that we remove, this illustrates a trade-off between noise reduction and loss of signal.
\begin{figure}
    \centering
    \includegraphics[width=0.5\linewidth]{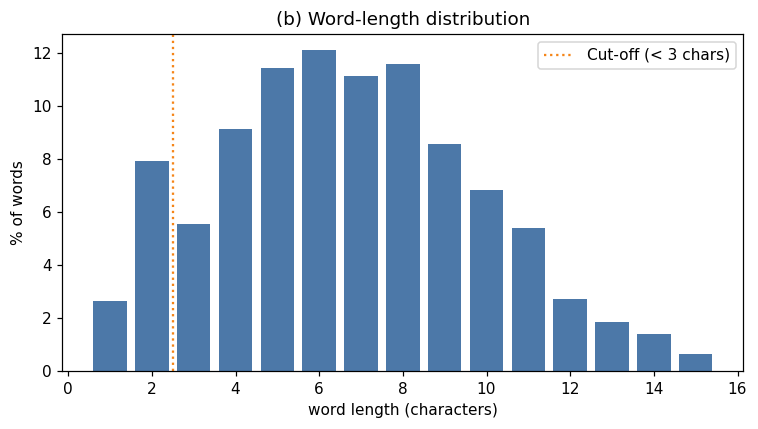}
    \caption{Distribution of the character counts of each of the words in the subjects and objects in the processed but unfiltered triples.}
    \label{fig:character_count}
\end{figure}

\begin{table}[h]
\centering
\caption{Examples of 1-character, 2-character and 3-character subjects and objects from raw triples. The chosen threshold to remove tokens was set at less than 3 characters, hence 1-character and 2-character tokens are removed.}
\label{tab:character_count}
\begin{tabular}{@{}ll@{}}
\toprule
                   & Examples                               \\ \midrule
1-character tokens & g, e, k, x, r, ., a, p, \%, v          \\
2-character tokens & of, in, al, we, et, ci, as, is, on, ar \\ 
3-character tokens & sum, dsc, low, set, the, and, liu, al., fit, all \\ \bottomrule
\end{tabular}
\end{table}

\paragraph{Frequency Threshold.} We now discuss the chosen frequency threshold, where we remove a triple if for either the subject or object all terms appear less than 5 times in the control corpus, as defined in Section~\ref{filtering}. We take this step to remove entities that are too rare. After all, we are interested in temporal trends and relations, hence entities need to have a minimum presence. This filtering step removes 5.5\% of all triples, indicating that most triples contain entities that are relatively present. To verify that our threshold is not too low, we randomly sample 20 of the removed entities, which are displayed in Table~\ref{tab:removed-entities}. We see that most removed entities are names, or niche terms such as \textit{socratic inquiry}. 

\begin{table}[h]
\centering
\caption{Twenty randomly sampled subject/object spans removed by the frequency filter
(every term has $f_{cc,t} < 5$ in the control corpus). They are predominantly author
names and other tokens absent from the general research vocabulary.}
\label{tab:removed-entities}
\small
\begin{tabular}{ll}
\toprule
\multicolumn{2}{c}{Removed subject/object spans} \\
\midrule
adam arvidsson & sangwoo chiheon sungwoong \\
gervet romoff batra parikh rabbat pineau & sicheng wang sabato marco siniscalchi chin-hui \\
sennrich alham fikri & yining hong \\
socratic inquiry & michael whalen \\
alex fiannaca & montral newtech \\
agnes luhtaru & derek xiuyu felix hohne ser-nam \\
christian hempelmann & aalbers haaren \\
benjamin piwowarski & bertossi salimi \\
laying fireplace & huaping zhong \\
iyengar singal & mustafa bozdag \\
\bottomrule
\end{tabular}
\end{table}

\subsubsection{Noun Stapling: SoftDICE threshold}\label{thresholdselection_nounstapling}

The noun-stapling step discussed in Section \ref{nounstapling} merges two compound nouns when their Soft-DICE similarity (Eq.~\ref{softdice}) is at least $0.85$. We chose this threshold to be conservative, so that only surface variants of the same entity are merged and distinct technologies are kept apart. This subsection shows more background on this choice.

Table~\ref{tab:dice-examples} lists example pairs at each value. We find that at values of $0.7$ and $0.75$, the entities can be significantly different, such as \textit{deep neural network watermark} and \textit{deep neural network modularization}. At values of $0.85$ and above, the compound nouns are highly similar, with only minor variations. In the end, we opted to put the threshold at $0.85$, as for $0.85$ there are still meaningful differences, for example \textit{proposed attacks} and \textit{proposed poisoning attacks}. We note that with this matching, there are still minor variations allowed between paired compound nouns. This is the balance we aim to strike between grouping identical signal together, while maintaining a high precision and reducing noise.

\begin{table}[t]
\centering
\caption{Example noun pairs at each achievable Soft-DICE value around the $0.85$
threshold, we allow for an allowed deviation of $0.02$ around the target value for each row. There were no samples with SoftDice values around $0.95$.
they are variants of the same entity.}
\label{tab:dice-examples}
\footnotesize
\begin{tabular}{@{}p{0.47\linewidth} p{0.47\linewidth}@{}}
\toprule
\multicolumn{2}{@{}l}{\textbf{SoftDICE $\approx 0.7$}} \\
learnable background token & learnable class-specific token \\
deferral loss function & variance-based loss function \\
current deep neural networks & hard dual deep neural networks \\
state training data & predictions training data \\
replaced causal model & causal imitative model \\
\addlinespace
\multicolumn{2}{@{}l}{\textbf{SoftDICE $\approx 0.75$}} \\
texture convolutional neural network & boosted convolutional neural network \\
deep neural network watermark & deep neural network modularization \\
analogy deep neural networks & poisoned deep neural networks \\
spatiotemporal gaussian process model & gaussian process density model \\
unknown deep neural network & iterative deep neural network \\
\addlinespace
\multicolumn{2}{@{}l}{\textbf{SoftDICE $\approx 0.80$}} \\
noise estimate & noise contrastive estimate \\
text-generation model & retrieval-augmented text-generation model \\
proposed attacks & proposed poisoning attacks \\
unique properties & unique chemical properties \\
model accuracy cost & accuracy cost \\
\addlinespace
\multicolumn{2}{@{}l}{\textbf{Soft-DICE $\approx 0.85$}} \\
framework multi-objective bayesian optimization & bayesian multi-objective optimization \\
radar waveform selection process & radar waveform selection \\
loss function employed & widely employed loss function \\
stepwise feature selection lasso & stepwise feature selection \\
dense neural network & three-layer dense neural network \\
\addlinespace
\multicolumn{2}{@{}l}{\textbf{SoftDICE $\approx 0.9$}} \\
using standard binary cross-entropy loss & using binary cross-entropy loss \\
gated recurrent unit relation-aware & relation-aware gated recurrent unit network \\
adaptive multi-hierarchical attention module & novel adaptive multi-hierarchical attention module \\
open-source proprietary large language & open-source proprietary large language models \\
autonomous robot bimanual manipulation behavior & robot bimanual manipulation behavior \\
\addlinespace
\multicolumn{2}{@{}l}{\textbf{SoftDICE $= 1.0$}} \\
classification accuracy rate & accuracy classification rate \\
shape pose parameters & pose shape parameters \\
visual linguistic modalities & linguistic visual modalities \\
student ensemble model & ensemble student model \\
charge prediction & prediction charge \\
\bottomrule
\end{tabular}
\end{table}

\subsection{Semantic noun-stapling}\label{app:semanticnounstapling}
Our work considers syntactic similarity for the noun stapling, as discussed in Section \ref{nounstapling}. However, this misses cases where two terms are semantically similar, but lexically different. For example, a missed abbreviation, such as \textit{llm} and \textit{large language model}. Or, alternatively, two different names for the same concept, such as \textit{prompt} and \textit{instruction}.

To test whether an embedding-based semantic normalization layer would help, we embedded the graph vocabulary with four sentence-embedding models spanning size and training regime, \texttt{all-MiniLM-L6-v2} (22M)~\footnote{https://huggingface.co/sentence-transformers/all-MiniLM-L6-v2}, \texttt{all-mpnet-base-v2} (110M)~\footnote{https://huggingface.co/sentence-transformers/all-mpnet-base-v2},
\texttt{allenai-specter} (110M)~\cite{specter}, and \texttt{google/embeddinggemma-300m} (300M)~\cite{embeddinggemma}.

The idea is that through semantic similarity, we would be able to link entities that are semantically identical but syntactically different. However, there are two concerns. First, one needs to verify that embedding models can accurately encode the information and identify whether two scientific entities are semantically equal or not, which research has shown to not always be the case~\cite{Wursch2023, alexandria}. Second, computing embedding similarities is computationally more costly than syntactic similarities. 

To assess the performance of embedding models, we manually curated three sets of entity pairs. The first category consists of semantically equal, but syntactically different entities (e.g. \textit{data annotation} and \textit{data labelling}). The second category consists of semantically different, but syntactically close entities  (e.g. \textit{logistic regression} and \textit{linear regression}). The last category consists of entities that are both syntactically and semantically distinct (e.g. \textit{convolutional neural network} and \textit{sentiment analysis}). We refer to these categories as \textit{Synonym}, \textit{Distinct} and \textit{Unrelated}. For semantic similarity to work, we need to be able to find a threshold where we both attain a low level of false negatives and false positives. The curated pairs are available in the repository.

\begin{figure}
    \centering
    \includegraphics[width=0.6\linewidth]{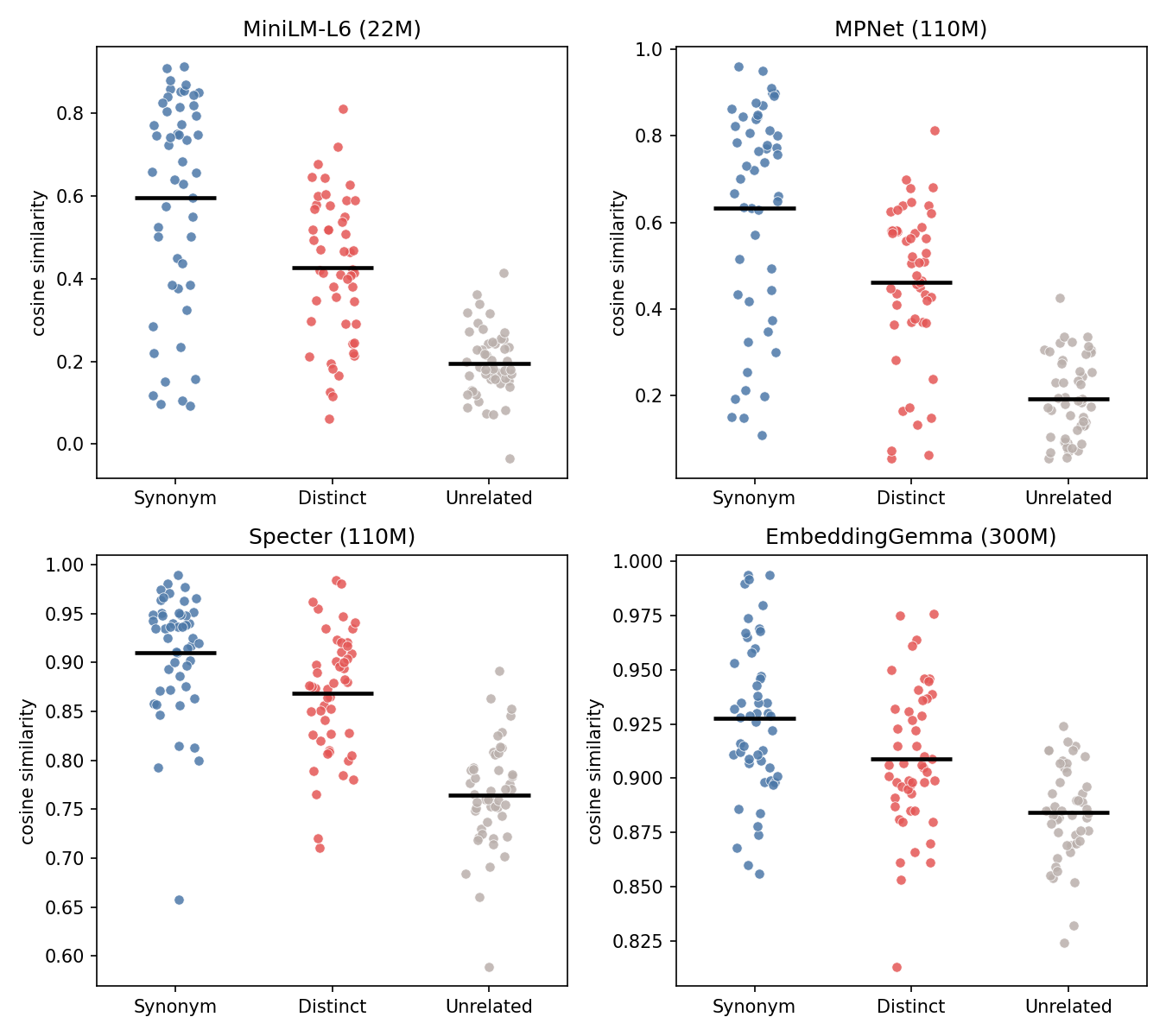}
    \caption{Cosine similarity of synonym, distinct and unrelated pairs for each embedding model.}
    \label{fig:embedding_appendix}
\end{figure}

Figure \ref{fig:embedding_appendix} shows the cosine similarities for the three categories of pairs, for each embedding model. We find that none of the embedding model cleanly manages to pair synonyms, while not falsely pairing distinct entities. We find that \texttt{MiniLM-L6} (22M) on average separates synonyms and distinct pairs better than \texttt{Specter} (110M). However, it also misses more synonym pairs, whereas \texttt{Specter} (110M) scores almost all synonym pairs above 0.80. Due to the computational cost of embedding similarity computation, and the inability of embedding models to cleanly pair entities, our work chooses to use syntactic similarity measures.

\subsection{Topic Modeling Methods}\label{app:topicmodelling}

In Section \ref{Key_term_extraction_methodology}, we describe how our work defines topics, to which the extracted triples are assigned later. To this end, we use KeyBERT, with \texttt{allenai/specter} as the embedding model. A natural question is how this approach compares to classical topic
models such as Latent Dirichlet Allocation (LDA)~\cite{lda} and
Non-negative Matrix Factorization (NMF)~\cite{nmf}. This appendix provides a comparison across these three methods, on the
$4{,}182$ LLM-related abstracts. We run LDA and NMF twice, once to fit with $20$ topics and once to fit with $50$ topics, on unigram, bigram and trigram features, so that they can also
surface multi-word terms.

\begin{figure}[h]
\centering
\begin{subfigure}{0.32\linewidth}
    \centering
    \includegraphics[width=\linewidth]{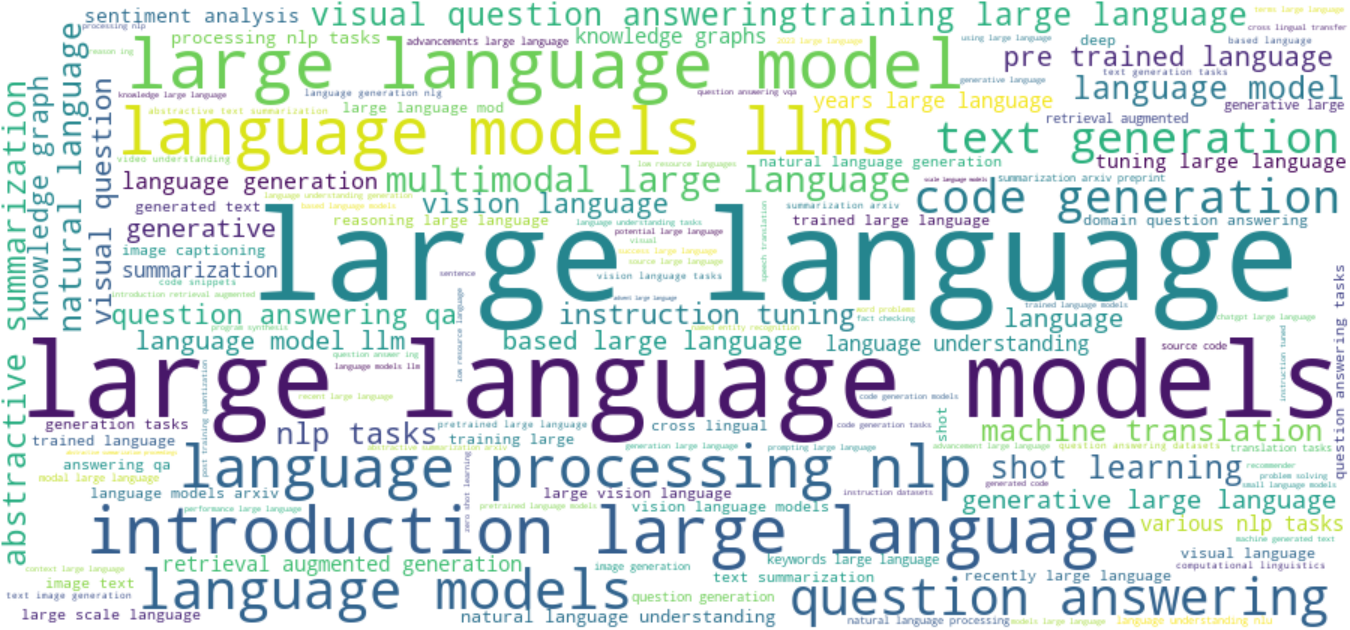}
    \caption{KeyBERT}
    \label{fig:wc-keybert}
\end{subfigure}
\hfill
\begin{subfigure}{0.32\linewidth}
    \centering
    \includegraphics[width=\linewidth]{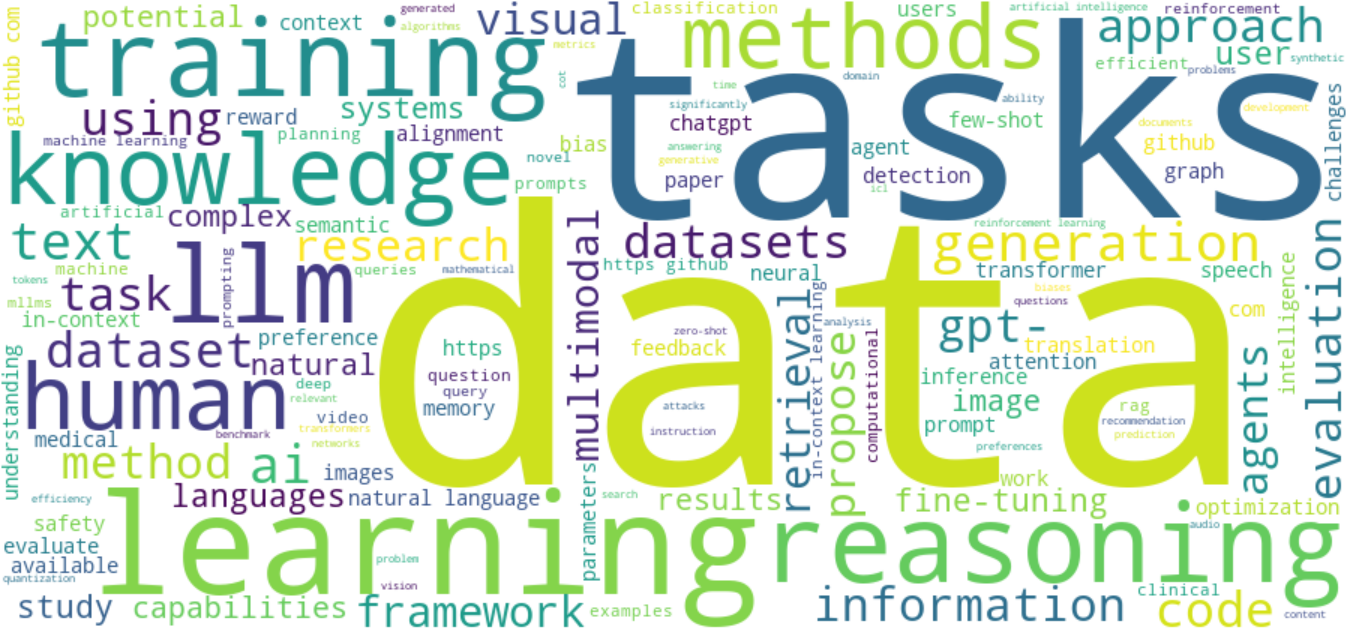}
    \caption{LDA (20 topics)}
    \label{fig:wc-lda}
\end{subfigure}
\hfill
\begin{subfigure}{0.32\linewidth}
    \centering
    \includegraphics[width=\linewidth]{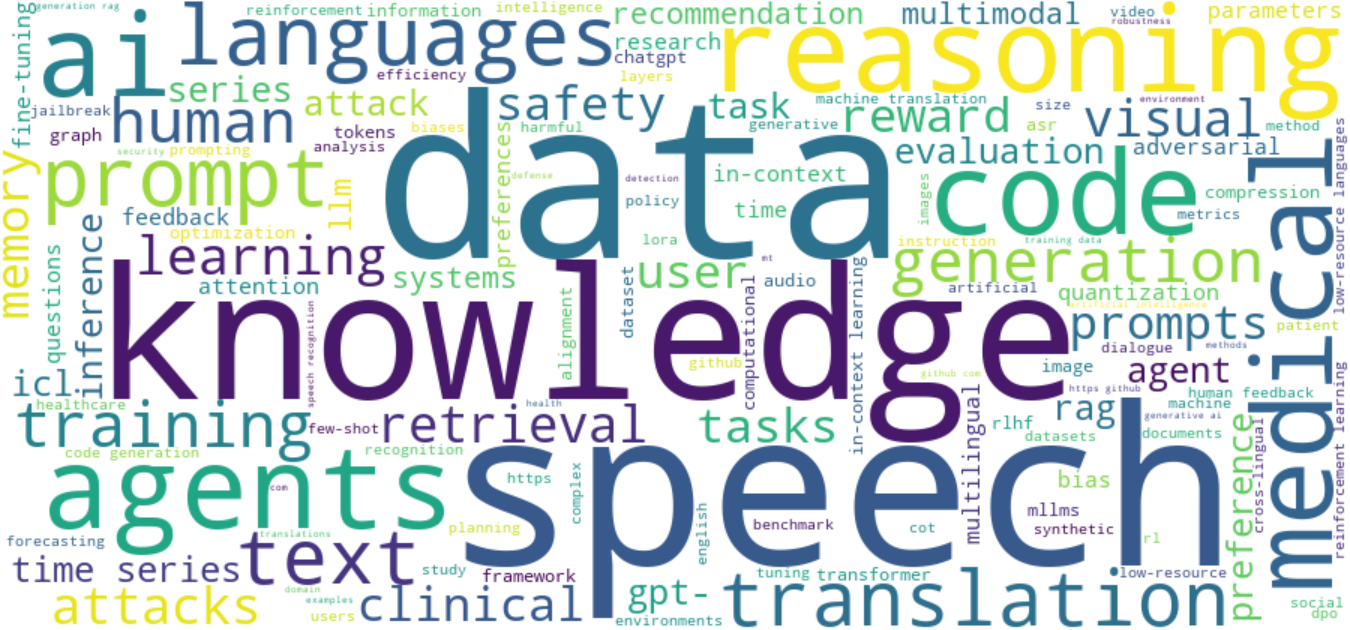}
    \caption{NMF (20 topics)}
    \label{fig:wc-nmf}
\end{subfigure}

\vspace{2pt}

\begin{subfigure}{0.32\linewidth}
    \centering
    \includegraphics[width=\linewidth]{images/wc_keybert.png}
    \caption{KeyBERT}
    \label{fig:wc-keybert50}
\end{subfigure}
\hfill
\begin{subfigure}{0.32\linewidth}
    \centering
    \includegraphics[width=\linewidth]{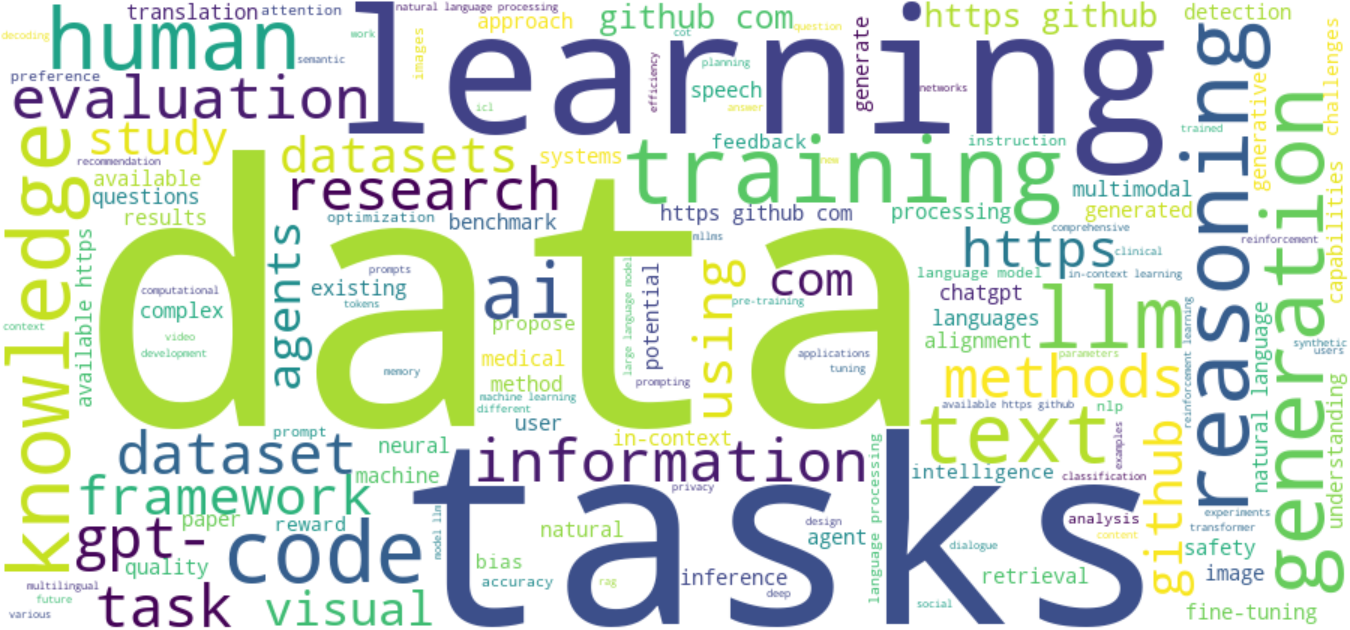}
    \caption{LDA (50 topics)}
    \label{fig:wc-lda50}
\end{subfigure}
\hfill
\begin{subfigure}{0.32\linewidth}
    \centering
    \includegraphics[width=\linewidth]{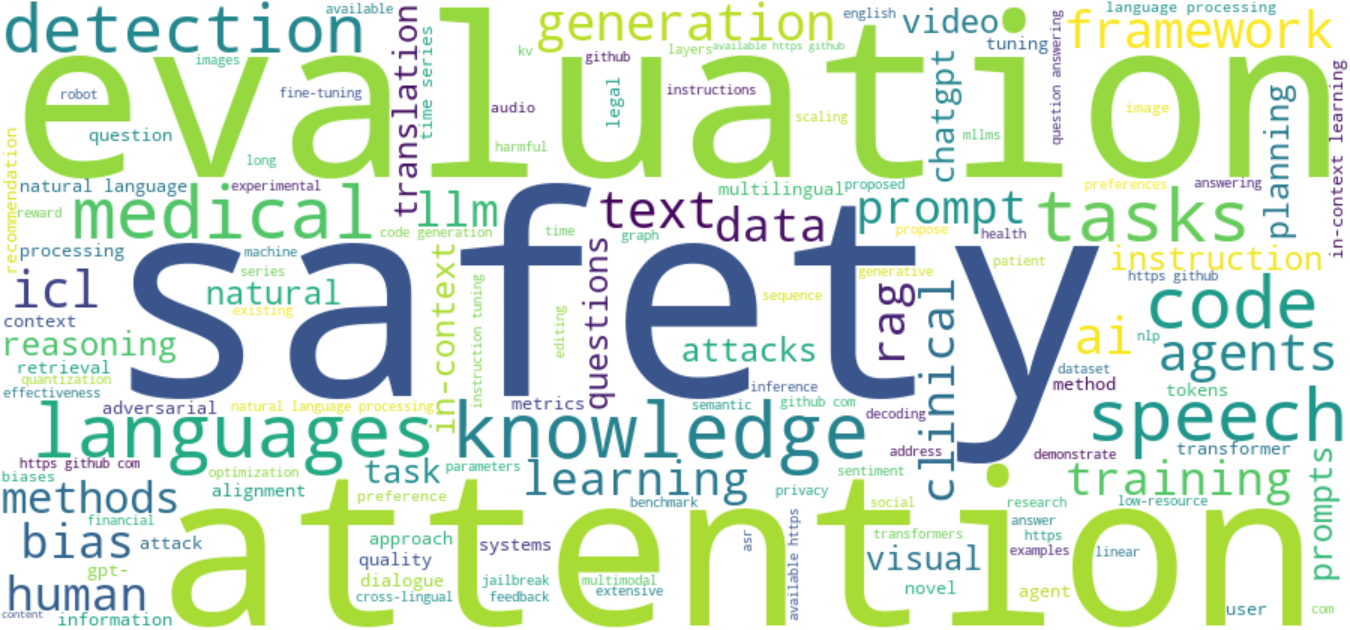}
    \caption{NMF (50 topics)}
    \label{fig:wc-nmf50}
\end{subfigure}

\caption{Word clouds of the terms surfaced by each method on the LLM-related abstracts. In the top row LDA and NMF at $20$ topics, in the bottom row at $50$ topics, with KeyBERT repeated for reference. KeyBERT surfaces
specific, multi-word technologies (e.g.\ \emph{visual question answering},
\emph{abstractive summarization}, \emph{instruction tuning}); LDA and NMF are dominated by broad, generic single words (\emph{tasks}, \emph{data},
\emph{learning}, \emph{knowledge}), and remain so even at $50$ topics.}
\label{fig:wordclouds}
\end{figure}

Figure \ref{fig:wordclouds} shows the terms surfaced by each method, with for LDA and NMF 20 topics at the top row, and 50 topics on the bottom row. We find that LDA and NMF mostly surface generic single terms, such as \textit{tasks} and \textit{data}. In contrast, KeyBERT surfaces more specific entities, such as \textit{large language model} and \textit{question answering}.  We validate this by randomly selecting 5 papers, and comparing the extracted entities from KeyBERT, LDA and NMF side by side. The results are displayed in Table \ref{tab:topic-per-paper}, where we again see that LDA and NMF lean towards single terms, that are relatively generic. In contrast, KeyBERT manages to extract more specific and fine-grained entities. For these reasons, we opt to use KeyBERT in our pipeline for topic identification. 

\begin{table}[t]
\centering
\caption{Extracted terms by KeyBERT, LDA and NMF, for five random papers. For LDA/NMF we use $20$ topics and unigram--trigram extraction.}
\label{tab:topic-per-paper}
\small
\begin{tabular}{@{}l p{0.80\linewidth}@{}}
\toprule
\multicolumn{2}{@{}l}{\textit{UQE: A Query Engine for Unstructured Databases~\cite{uqe}}} \\
KeyBERT & unstructured data analytics, structured data, query execution, query language uql, query engine \\
LDA & data, retrieval, methods, natural, propose, new \\
NMF & data, retrieval, queries, query, natural language, methods \\
\addlinespace
\multicolumn{2}{@{}l}{\textit{LLM Agents Improve Semantic Code Search~\cite{llmagents}}} \\
KeyBERT & semantic code search, code retrieval, code retrieval systems, enhance code retrieval \\
LDA & retrieval, information, context, rag, code, generation \\
NMF & retrieval, rag, generation rag, code, information \\
\addlinespace
\multicolumn{2}{@{}l}{\textit{Evaluating the Logical Reasoning Ability of ChatGPT and GPT-4~\cite{liu2023}}} \\
KeyBERT & chatgpt performs significantly, generative pretrained transformer, comprehension natural language \\
LDA & reasoning, tasks, gpt, natural language, benchmark \\
NMF & reasoning, reasoning tasks, logical, gpt, ability \\
\addlinespace
\multicolumn{2}{@{}l}{\textit{DEE: Dual-stage Explainable Evaluation Method for Text Generation~\cite{dee}}} \\
KeyBERT & text generation, machine generated texts, generated texts, quality text generation \\
LDA & text, generation, evaluation, generated, dataset, systems \\
NMF & text, evaluation, human, generation, quality \\
\addlinespace
\multicolumn{2}{@{}l}{\textit{End-to-End Ontology Learning with Large Language Models~\cite{lo2024}}} \\
KeyBERT & ontology learning, ontology finetuning, target ontology finetuning, partial ontology learning \\
LDA & code, datasets, methods, github, https, learning \\
NMF & code, knowledge, github, https, github com \\
\bottomrule
\end{tabular}
\end{table}

\subsection{Predicate Analysis} \label{app:predicate_analysis}

In the case studies discussed in Sections \ref{casestudy1} and \ref{casestudy2}, we focus on the subjects and objects of the triples. In this appendix, we perform a high-level analysis on the predicates of the triples, to investigate whether they provide us with additional insights. 

\begin{figure}[t]
\centering
\includegraphics[width=0.75\linewidth]{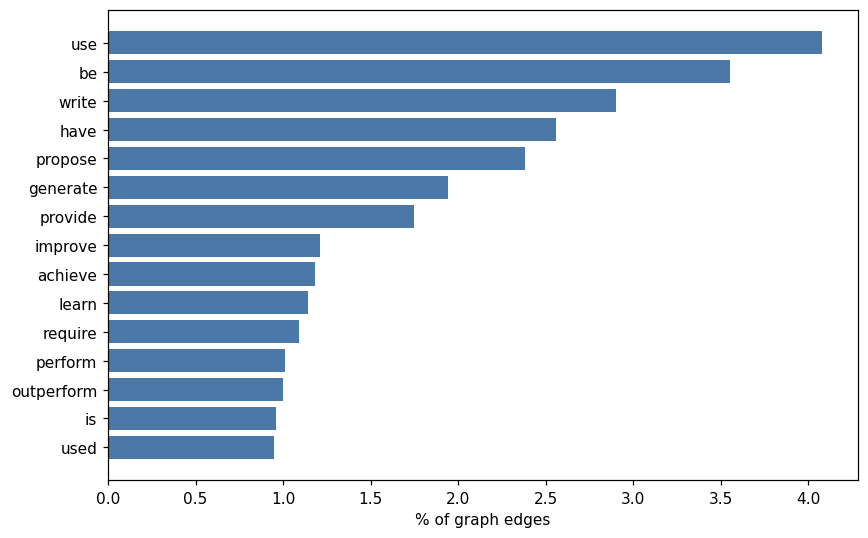}
\caption{The most frequent predicate head verbs, as a share of all graph edges.
The distribution is dominated by generic discourse and usage verbs.}
\label{fig:predicate-verbs}
\end{figure}

\paragraph{Most Frequent Predicates.} Figure~\ref{fig:predicate-verbs} shows
the most frequent predicate verbs across the $2.1$M final graph edges. The
distribution is dominated by generic scientific-discourse and usage verbs
(\emph{use}, \emph{be}, \emph{write}, \emph{have}, \emph{propose},
\emph{provide}): the ten most frequent head verbs already account for $22.7\%$ of
all edges and describe how authors write about their work rather than a specific
relation between two technologies. 

\paragraph{Predicate Categorization.} To characterize the predicate space, we assign each head verb to its lexicographic class in
WordNet~\footnote{https://wordnet.princeton.edu/documentation/lexnames5wn}~\cite{wordnet}. Because many common verbs are highly polysemous, and their single most
frequent WordNet sense can be misleading (for example the dominant sense of
\emph{play} would place ``plays a role'' in \emph{competition}), we use a
sense-agreement filter: a verb is assigned to a class only when a majority of its
verb senses share that lexname, and is otherwise left \emph{ambiguous}. This
labels $59.9\%$ of edges (Table~\ref{tab:predicate-wordnet}). The predicates that
express a directed technical relation, such as\emph{change}, \emph{creation}, and
\emph{competition}, are a minority. The larger classes
(\emph{stative}, \emph{cognition}, \emph{communication}) are reporting and
description verbs. This aligns with Figure \ref{fig:predicate-verbs}, which shows that the most common verbs are generic. 

However, for a qualitative evaluation, one can still consider the smaller informative predicate classes, such as \emph{competition}. Here, for example, we see that it is dominated by the predicate \emph{outperform} ($21{,}206$ edges). These predicates encode \emph{technology
succession}: a triple $(\textrm{X}, \textrm{outperform}, \textrm{Y})$ states that
technology X supersedes technology Y on some task. Aggregating the objects of
\emph{outperform} therefore traces which established technologies are most
frequently reported as surpassed (Table~\ref{tab:outperform-year}). The result is
intuitive: recent, strong baselines such as \emph{ChatGPT}, \emph{multilingual
BERT}, \emph{in-context learning}, \emph{fine-tuning} and \emph{federated
learning} are the technologies most often outperformed. Individual edges are
equally interpretable, for example \emph{deep learning} $\rightarrow$
\emph{traditional machine learning-based methods}, and \emph{structured prompt
tuning} $\rightarrow$ \emph{standard prompt tuning}.

\begin{table}[t]
\centering
\caption{Predicate head verbs of the graph edges, categorized by WordNet
lexicographic class with a sense-agreement filter. Ambiguous covers highly
polysemous verbs with no majority class.}
\label{tab:predicate-wordnet}
\begin{tabular}{lrr}
\toprule
WordNet class & Edges & \% \\
\midrule
ambiguous & 847{,}088 & 40.1 \\
creation & 215{,}403 & 10.2 \\
stative & 211{,}616 & 10.0 \\
cognition & 182{,}463 & 8.6 \\
change & 165{,}067 & 7.8 \\
consumption & 149{,}785 & 7.1 \\
communication & 141{,}027 & 6.7 \\
social & 77{,}274 & 3.7 \\
possession & 37{,}801 & 1.8 \\
competition & 31{,}379 & 1.5 \\
contact & 26{,}452 & 1.3 \\
perception & 18{,}023 & 0.9 \\
other (motion, body, emotion, weather) & 9{,}080 & 0.4 \\
\bottomrule
\end{tabular}
\end{table}

\begin{table}[t]
\centering
\caption{Top-5 technologies on each side of the predicate \emph{outperform}, by
year. The lists track the shift from deep learning, to BERT and prompt/fine-tuning,
to ChatGPT and retrieval-augmented generation.}
\label{tab:outperform-year}
\footnotesize
\begin{tabular}{@{}p{0.46\linewidth} p{0.46\linewidth}@{}}
\toprule
Winner (X outperforms $\cdot$) & Surpassed ($\cdot$ outperformed by X) \\
\midrule
\multicolumn{2}{c}{\textbf{2018}} \\
deep learning & single-task learning \\
gated path planning network & value iteration network \\
biomedical translation model & multi-task learning \\
self-paced partial-label learning & online learning algorithms \\
semantically conditioned variational autoencoder & deep learning methods \\
\addlinespace
\multicolumn{2}{c}{\textbf{2020}} \\
deep learning & multilingual bert \\
multilingual bert & deep learning models \\
federated learning & multilingual model \\
multi-task learning & model-agnostic meta-learning \\
learning to retrieve & multilingual models \\
\addlinespace
\multicolumn{2}{c}{\textbf{2022}} \\
prompt tuning & fine-tuning \\
multilingual bert & multilingual bert \\
federated learning & deep learning models \\
deep learning models & multilingual models \\
deep learning & strong baselines \\
\addlinespace
\multicolumn{2}{c}{\textbf{2024}} \\
chatgpt & chatgpt \\
federated learning & in-context learning \\
in-context learning & zero-shot \\
retrieval-augmented generation & deep learning models \\
deep learning & zero-shot prompting \\
\bottomrule
\end{tabular}
\end{table}